\documentclass{article}

\usepackage{listings}
\usepackage[dvipsnames]{xcolor}

\definecolor{codegreen}{rgb}{0,0.6,0}
\definecolor{codegray}{rgb}{0.5,0.5,0.5}
\definecolor{codepurple}{rgb}{0.58,0,0.82}
\definecolor{backcolour}{rgb}{0.95,0.95,0.92}

\lstdefinestyle{mystyle}{
    backgroundcolor=\color{backcolour},   
    commentstyle=\color{codegreen},
    keywordstyle=\color{magenta},
    numberstyle=\tiny\color{codegray},
    stringstyle=\color{codepurple},
    basicstyle=\ttfamily\footnotesize,
    breakatwhitespace=false,
    breaklines=true,         
    captionpos=b,
    keepspaces=true,
    numbers=left,
    numbersep=5pt,
    showspaces=false,
    showstringspaces=false,
    showtabs=false,
    tabsize=1
}
 
\usepackage{amsmath,amsthm,amsfonts,amssymb}
\usepackage{bm}

\usepackage{tablefootnote}
\usepackage{multirow}
\usepackage{hhline}
\usepackage{graphicx}

\usepackage{algorithm,algorithmic}
\renewcommand{\algorithmiccomment}[1]{\ \ \textcolor{OliveGreen}{\# #1}}

\usepackage{caption}
\usepackage{subcaption}
\usepackage{url}

\DeclareMathOperator*{\argmax}{\mathrm{arg\,max}}
\newcommand{\bmX}{\bm{X}}\newcommand{\bmg}{\bm{g}}\newcommand{\bmh}{\bm{h}}\newcommand{\bmx}{\bm{x}}

\newtheorem{theorem}{Theorem}
\newtheorem{definition}{Definition}

\usepackage{microtype}
\usepackage{graphicx}
\usepackage{booktabs} 

\usepackage{hyperref}

\usepackage[accepted]{icml2020}

\begin{document}

\twocolumn[
\icmltitle{Do We Need Zero Training Loss After Achieving Zero Training Error?}



\icmlsetsymbol{equal}{*}

\begin{icmlauthorlist}
\icmlauthor{Takashi Ishida}{ut,ri}
\icmlauthor{Ikko Yamane}{ut}
\icmlauthor{Tomoya Sakai}{ne}
\icmlauthor{Gang Niu}{ri}
\icmlauthor{Masashi Sugiyama}{ri,ut}
\end{icmlauthorlist}

\icmlaffiliation{ut}{The University of Tokyo}
\icmlaffiliation{ri}{RIKEN}
\icmlaffiliation{ne}{NEC Corporation}

\icmlcorrespondingauthor{Takashi Ishida}{ishida@ms.k.u-tokyo.ac.jp}

\icmlkeywords{Machine Learning, ICML}

\vskip 0.3in
]



\printAffiliationsAndNotice{}  

\begin{abstract}
Overparameterized deep networks have the capacity to memorize training data with zero \emph{training error}.
Even after memorization, the \emph{training loss} continues to approach zero, making the model overconfident and the test performance degraded.
Since existing regularizers do not directly aim to avoid zero training loss, 
it is hard to tune their hyperparameters in order to maintain a fixed/preset level of training loss.
We propose a direct solution called \emph{flooding} that intentionally prevents further reduction of the training loss when it reaches a reasonably small value, which we call the \emph{flood level}.
Our approach makes the loss float around the flood level by doing mini-batched gradient descent as usual but gradient ascent if the training loss is below the flood level.
This can be implemented with one line of code and is compatible with any stochastic optimizer and other regularizers.
With flooding, the model will continue to ``random walk'' with the same non-zero training loss, and we expect it to drift into an area with a flat loss landscape that leads to better generalization.
We experimentally show that flooding improves performance and, as a byproduct, induces a double descent curve of the test loss.
\end{abstract}

\section{Introduction}
\label{sec:introduction}
\begin{figure}[ht]
\centering
\subcaptionbox{w/o Flooding \label{fig:overfittings-a}}{\includegraphics[width=\columnwidth*21/48]{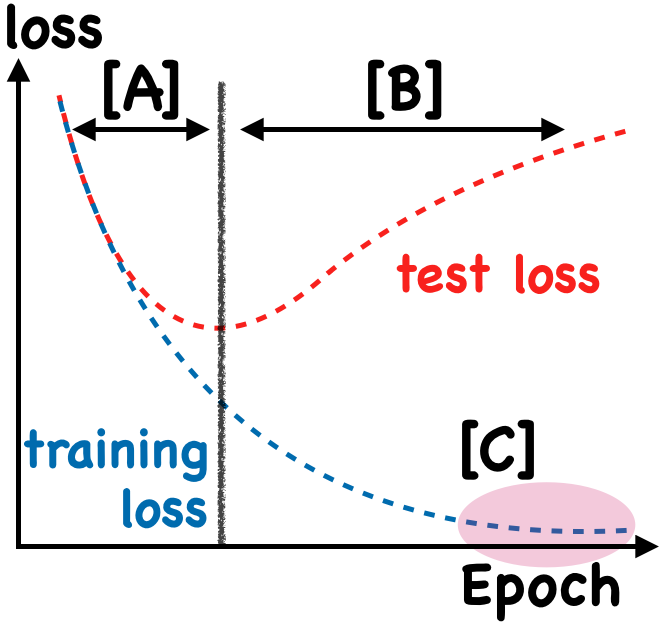}}
\hspace{1.5mm}
\subcaptionbox{w/ Flooding\label{fig:overfittings-b}}{\includegraphics[width=\columnwidth*21/48]{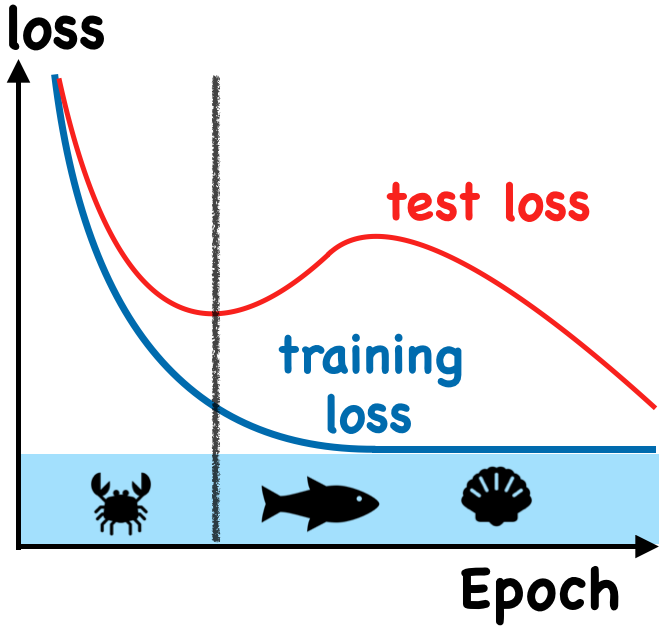}}\\
\subcaptionbox{C10 w/o Flooding\label{fig:overfittings-c}}{\includegraphics[width=\columnwidth*11/24]{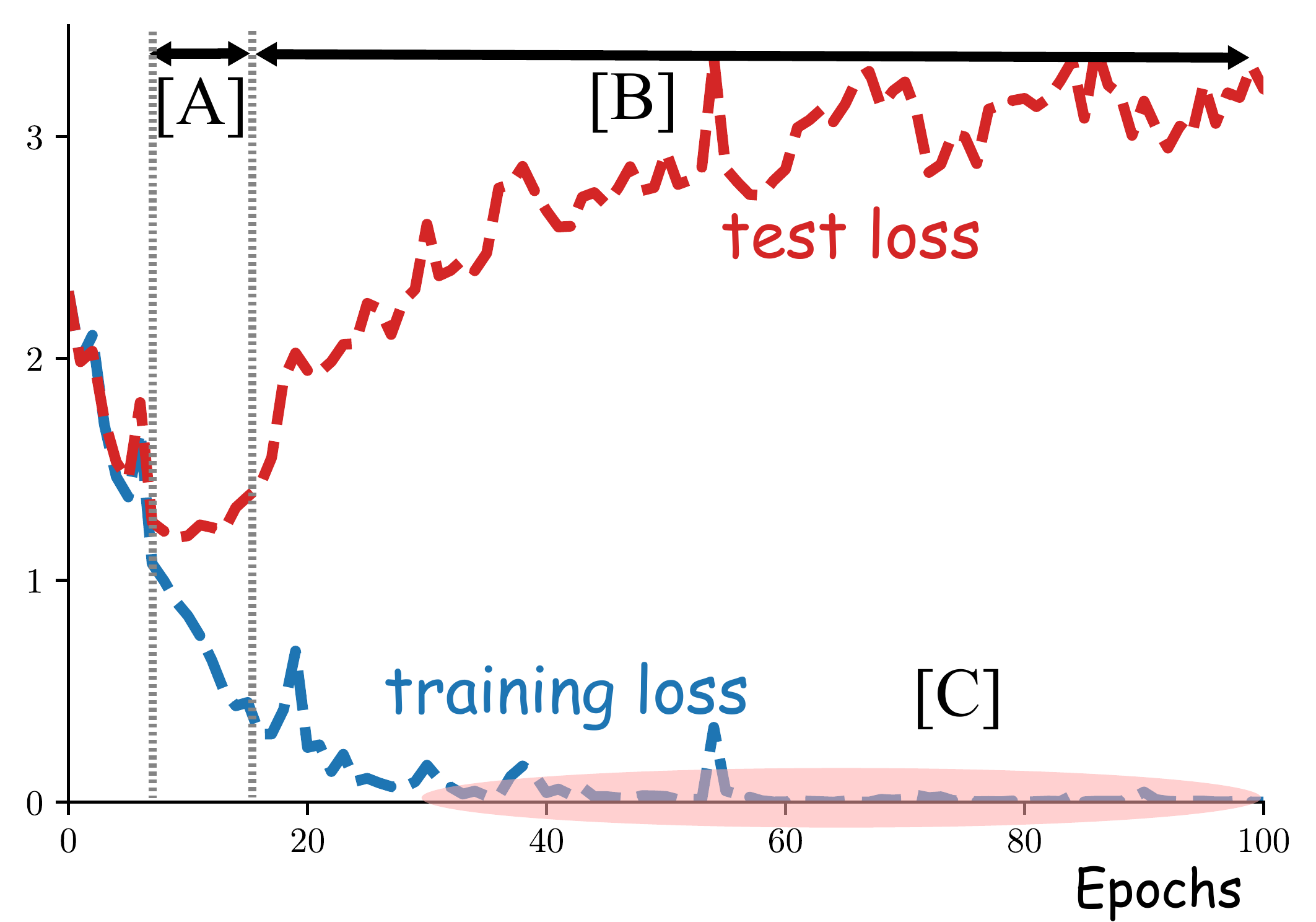}}
\subcaptionbox{C10 w/ Flooding\label{fig:overfittings-d}}{\includegraphics[width=\columnwidth*11/24]{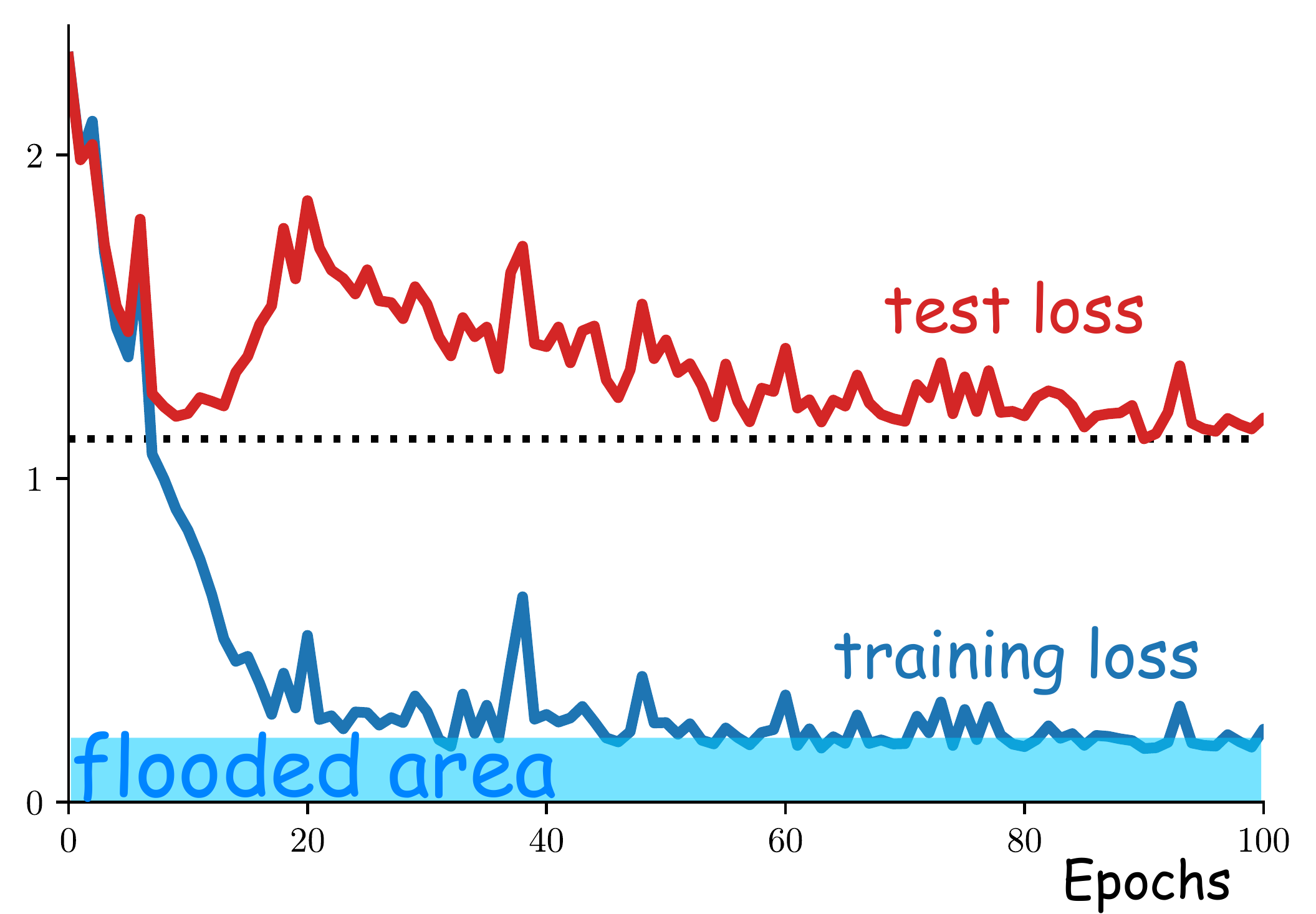}}
\caption{
(a) shows 3 different concepts related to overfitting.
[A] shows the generalization gap increases, while the training and test losses decrease.
[B] also shows the increasing gap, but the test loss starts to rise.
[C] shows the training loss becoming (near-)zero.
We avoid [C] by \emph{flooding} the bottom area, visualized in (b), which forces the training loss to stay around a constant.
This leads to a decreasing test loss once again.
We confirm these claims in experiments with CIFAR-10 shown in (c)--(d).
}
\label{fig:overfittings}
\vskip -0.25in
\end{figure}

``Overfitting'' is one of the biggest interests and concerns in the machine learning community \cite{ng1997icml,caruana2000neurips, belkin2018neurips, roelofs2019neurips,wepachowski2019neurips}.
One way of identifying overfitting is to see whether the generalization gap, the test minus the training loss, is increasing or not \cite{goodfellow2016book}.
We can further decompose the situation of the generalization gap increasing into two stages:
The first stage is when both the training and test losses are decreasing, but the training loss is decreasing faster than the test loss ([A] in Fig.~\ref{fig:overfittings-a}.)
The next stage is when the training loss is decreasing, but the test loss is increasing.

Within stage [B], after learning for even more epochs, the training loss will continue to decrease and may become (near-)zero.
This is shown as [C] in Fig.~\ref{fig:overfittings-a}.
If we continue training even after the model has memorized \cite{zhang2017iclr,arpit2017icml,belkin2018neurips} the training data completely with zero error, the training loss can easily become (near-)zero especially with overparametrized models.
Recent works on overparametrization and double descent curves \cite{belkin2019pnas,nakkiran2020iclr} have shown that learning until zero training error is meaningful to achieve a lower generalization error.
However, whether zero training \emph{loss} is necessary after achieving zero training \emph{error} remains an open issue.

In this paper, we propose a method to make the training loss \emph{float} around a small constant value, in order to prevent the training loss from approaching zero.
This is analogous to \emph{flooding} the bottom area with water, and we refer to the constant value as the \emph{flood level}.
Note that even if we add flooding, we can still memorize the training data.
Our proposal only forces the training \emph{loss} to become positive, which does not necessarily mean the training \emph{error} will become positive, as long as the flood level is not too large.
The idea of flooding is shown in Fig.~\ref{fig:overfittings-b},
and we show learning curves before and after flooding with a benchmark dataset in Fig.~\ref{fig:overfittings-c} and Fig.~\ref{fig:overfittings-d}.\footnote{For the details of these experiments, see Appendix~\ref{appsec:learning_curves}.}

\paragraph{Algorithm and implementation}
Our algorithm of flooding is surprisingly simple.
If the original learning objective is $J$, the proposed modified learning objective $\widetilde{J}$ with flooding is
\begin{equation}
\label{eq:flooding1}
    \widetilde{J}(\bm{\theta}) = |J(\bm{\theta}) - b| + b,
\end{equation}
where $b>0$ is the flood level specified by the user, and $\bm{\theta}$ is the model parameter.\footnote{
Adding back $b$ will not affect the gradient but will ensure $\widetilde{J}(\bm{\theta}) = J(\bm{\theta})$ when $J(\bm{\theta}) > b$.
}

The gradient of $\tilde{J}$ w.r.t.\@ $\bm{\theta}$ will point in the same direction as that of $J(\bm{\theta})$ when $J(\bm{\theta})>b$ but in the opposite direction when $J(\bm{\theta})<b$.
This means that when the learning objective is above the flood level, there is a ``gravity'' effect with gradient \textit{descent}, but when the learning objective is below the flood level, there is a ``buoyancy''  effect with gradient \textit{ascent}.
In practice, this will be performed with a mini-batch and will be compatible with any stochastic optimizers.
It can also be used along with other  regularization methods.

During flooding, the training loss will repeat going below and above the flood level.  The model will continue to ``random walk'' with the same non-zero training loss, and we expect it to drift into an area with a flat loss landscape that leads to better generalization \cite{schmidhuber1997neuco,chaudhari2017iclr,keskar2017iclr,li2018neurips}.
In experiments, we show that during this period of random walk, there is an increase in flatness of the loss function (See Section~\ref{subsec:flatness}).

This modification can be incorporated into existing machine learning code easily:
Add one line of code for Eq.~\eqref{eq:flooding1} after evaluating the original objective function $J(\bm{\theta})$.
A minimal working example with a mini-batch in PyTorch \cite{paszke2019neurips} is demonstrated below to show the additional one line of code:
\begin{lstlisting}[language=Python]
outputs = model(inputs)
loss = criterion(outputs, labels)
flood = (loss-b).abs()+b  # This is it!
optimizer.zero_grad()
flood.backward()
optimizer.step()
\end{lstlisting}

It may be hard to set the flood level without expert knowledge on the domain or task.
We can circumvent this situation easily by treating the flood level as a hyper-parameter.
We may exhaustively evaluate the accuracy for the predefined hyper-parameter candidates with a validation dataset, which can be performed in parallel.

\paragraph{Previous regularization methods}
Many previous regularization methods also aim at avoiding \emph{training too much} in various ways including restricting the parameter norm to become small by decaying the parameter weights \cite{hanson1988nips}, raising the difficulty of training by dropping activations of neural networks \cite{srivastava2014jmlr}, avoiding the model to output a hard label by smoothing the training labels \cite{szegedy2016cvpr}, and simply stopping training at an earlier phase \cite{morgan1990nips}.
These methods can be considered as indirect ways to control the training loss, by also introducing additional assumptions such as the optimal model parameters are close to zero.
Although making the regularization effect stronger would make it harder for the training loss to approach zero, it is still hard to maintain the right level of training loss till the end of training.
In fact, for overparametrized deep networks, applying small regularization would not stop the training loss becoming (near-)zero, making it even harder to choose a hyper-parameter that corresponds to a specific level of loss.

Flooding, on the other hand, is a direct solution to the issue that the training loss becomes (near-)zero.
Flooding intentionally prevents further reduction of the training loss when it reaches a reasonably small value, and the flood level corresponds to the level of training loss that the user wants to keep.

\begin{table*}[t]
\caption{
Conceptual comparisons of various regularizers.
``Indep.''/``tr.'' stands for ``independent''/``training''
and $\checkmark$/$\times$ stands for yes/no.
}
\label{tb:regularization}
\vskip 0.0in
\centering
\begin{footnotesize}
\begin{tabular}{l|ccccl}
\toprule
Regularization and other methods & \begin{tabular}{@{}c@{}}Target\\tr. loss\end{tabular} &
\begin{tabular}{@{}c@{}}Domain\\indep.\end{tabular} & \begin{tabular}{@{}c@{}}Task\\indep.\end{tabular} & \begin{tabular}{@{}c@{}}Model\\indep.\end{tabular}  & Main assumption\\
\midrule
$\ell_2$ regularization \cite{tikhonov1943sssr} & $\times$ & $\checkmark$ & $\checkmark$ & $\checkmark$ & Optimal model params are close to 0\\
Weight decay \cite{hanson1988nips} & $\times$ & $\checkmark$ & $\checkmark$ & $\checkmark$ & Optimal model params are close to 0\\
Early stopping \cite{morgan1990nips} & $\times$ & $\checkmark$ & $\checkmark$ & $\checkmark$ & Overfitting occurs in later epochs\\
$\ell_1$ regularization \cite{tibshirani1996jrss} & $\times$ & $\checkmark$ & $\checkmark$ & $\checkmark$ & Optimal model has to be sparse\\
Dropout \cite{srivastava2014jmlr} & $\times$ & $\checkmark$ & $\checkmark$ & $\times$ & Existence of complex co-adaptations\\
Batch normalization \cite{ioffe2015icml} & $\times$ & $\checkmark$ & $\checkmark$ & $\times$ & Existence of internal covariate shift\\
Label smoothing \cite{szegedy2016cvpr} & $\times$ & $\checkmark$ & $\times$ & $\checkmark$ & True posterior is not a one-hot vector\\
Mixup \cite{zhang2018iclr} & $\times$ & $\times$ & $\times$ & $\checkmark$ & Linear relationship between $\bmx$ and $y$\\
Image augment. \cite{shorten2019jbg} & $\times$ & $\times$ & $\checkmark$ & $\checkmark$ & Input is invariant to the translations\\\midrule
Flooding (proposed method) & $\checkmark$ & $\checkmark$ & $\checkmark$ & $\checkmark$ & Learning until zero loss is harmful\\
\bottomrule
\end{tabular}
\end{footnotesize}
\vskip -0.1in
\end{table*}

\section{Backgrounds}
\label{sec:related_works}
In this section, we review regularization methods (summarized in Table~\ref{tb:regularization}), recent works on overparametrization and double descent curves, and the area of weakly supervised learning where similar techniques to flooding have been explored.

\subsection{Regularization Methods}
\label{sec:related_regularization}
The name ``regularization'' dates back to at least Tikhonov regularization for the ill-posed linear least-squares problem~\cite{tikhonov1943sssr, tikhonov1977book}.
One example is to modify $\bmX^\top\bmX$ (where $\bmX$ is the design matrix) to become ``regular'' by adding a term to the objective function.
$\ell_2$ regularization is a generalization of the above example and can be applied to non-linear models.
These methods implicitly assume that the optimal model parameters are close to zero.

It is known that weight decay \cite{hanson1988nips}, dropout \cite{srivastava2014jmlr}, and early stopping \cite{morgan1990nips} are equivalent to $\ell_2$ regularization under certain conditions \cite{loschilov2019iclr,bishop1995icann, goodfellow2016book, wager2013nips}, implying that they have similar assumptions on the optimal model parameters.
There are other penalties based on different assumptions such as the $\ell_1$ regularization \cite{tibshirani1996jrss} based on the sparsity assumption that the optimal model has only a few non-zero parameters.

Modern machine learning tasks are applied to complex problems where the optimal model parameters are not necessarily close to zero or are not sparse, and it would be ideal if we can properly add regularization effects to the optimization stage without such assumptions.
Our proposed method does not have assumptions on the optimal model parameters and can be useful for more complex problems.

More recently, ``regularization'' has further evolved to a more general meaning including various methods that alleviate overfitting but do not necessarily have a step to regularize a singular matrix or add a regularization term to the objective function.
For example, \citet{goodfellow2016book} defines regularization as ``any modification we make to a learning algorithm that is intended to reduce its generalization error but not its training error.''
In this paper, we adopt this broader meaning of ``regularization.''

Examples of the more general regularization category include mixup \cite{zhang2018iclr} and data augmentation methods like cropping, flipping, and adjusting brightness or sharpness \cite{shorten2019jbg}.
These methods have been adopted in many state-of-the-art methods \cite{verma2019ijcal, Berthelot2019neurips,kolesnikov2020arxiv} and are becoming essential regularization tools for developing new systems.
However, these regularization methods have the drawback of being domain-specific: They are designed for the vision domain and require some efforts when applying to other domains \cite{guo2019arxiv,thulasidasan2019neurips}.
Other regularizers such as label smoothing \cite{szegedy2016cvpr} is used for problems with class labels and harder to use with regression or ranking, meaning that they are task-specific.
Batch normalization \cite{ioffe2015icml} and dropout \cite{srivastava2014jmlr}
are designed for neural networks and are model-specific.

Although these regularization methods---both the \emph{special} and \emph{general} ones---already work well in practice and have become the de facto standard tools \cite{bishop2011book, goodfellow2016book}, we provide an alternative which is even more general in the sense that it is domain-, task-, and model-independent.

That being said, we want to emphasize that the most important difference between flooding and other regularization methods is whether it is possible to target a specific level of training loss other than zero.
While flooding allows the user to choose the level of training loss directly, it is hard to achieve this with other regularizers.

\subsection{Double Descent Curves with Overparametrization}
Recently, there has been increasing attention on the phenomenon of ``double descent,'' named by \citet{belkin2019pnas}, to explain the two regimes of deep learning:
The first one (underparametrized regime) occurs where the model complexity is small compared to the number of samples, and the test error as a function of model complexity decreases with low model complexity but starts to increase after the model complexity is large enough.
This follows the classical view of machine learning that excessive complexity leads to poor generalization.
The second one (overparametrized regime) occurs when an even larger model complexity is considered.
Then increasing the complexity only decreases test error, which leads to a double descent shape.
The phase of decreasing test error often occurs after the training error becomes zero.
This follows the modern view of machine learning that bigger models lead to better generalization.\footnote{\url{https://www.eff.org/ai/metrics}}

As far as we know, the discovery of double descent curves dates back to at least \citet{krogh1992neurips}, where they theoretically showed the double descent phenomenon under a linear regression setup.
Recent works \cite{belkin2019pnas, nakkiran2020iclr} have shown empirically that a similar phenomenon can be observed with deep learning methods.
\citet{nakkiran2020iclr} observed that the double descent curves for the \emph{test error} can be shown not only as a function of model complexity, but also as a function of the epoch number.

To the best of our knowledge, the epoch-wise double descent curve was not observed for the \emph{test loss} before but was observed in our experiments after using flooding with only about 100 epochs.
Investigating the connection between epoch-wise double descent curves for the test loss and previous double descent curves \cite{krogh1992neurips, belkin2019pnas, nakkiran2020iclr} is out of the scope of this paper but is an important future direction.

\subsection{Avoiding Over-Minimization of the Empirical Risk}
It is commonly observed that the empirical risk goes below zero, and it causes overfitting \cite{kiryo17nips} in \emph{weakly supervised learning}
when an equivalent form of the risk expressed with the given weak supervision is alternatively used \cite{natarajan2013nips,jesus14ecml,christo14neurips,christo15icml,patrini17cvpr,rooyen2018jmlr}.
\citet{kiryo17nips} proposed a gradient ascent technique to keep the empirical risk non-negative.
This idea has been generalized and applied to other weakly supervised settings \cite{han2020icml, ishida2019icml,lu2020aistats}.

Although we also set a lower bound on the empirical risk, the motivation is different:
First, while \citet{kiryo17nips} and others aim to fix the negative empirical risk to become non-negative,
our original empirical risk is already non-negative. 
Instead, we are aiming to sink the original empirical risk by modifying it with a \emph{positive} lower bound.
Second, the problem settings are different.
Weakly supervised learning methods require certain loss corrections or sample corrections \cite{han2020icml} before the non-negative correction, but we work on the original empirical risk without any setting-specific modifications.

Early stopping \cite{morgan1990nips} may be a naive solution to this problem
where the empirical risk becomes too small.
However, performance of early stopping highly depends on the training dynamics and is sensitive to the randomness in the optimization method and mini-batch sampling.
This suggests that early stopping at the optimal epoch in terms of a single training path does not necessarily perform well in another round of training.
This makes it difficult to use hyper-parameter selection techniques such as cross-validation that requires re-training a model more than once.
In our experiments, we will demonstrate how flooding performs even better than early stopping.

\section{Flooding: How to Avoid Zero Training Loss}
\label{sec:proposed_method}
In this section, we propose our regularization method, \emph{flooding}.
Note that this section and the following sections only consider multi-class classification for simplicity.

\subsection{Preliminaries}
\label{sec:preliminaries}
Consider input variable $\bmx \in \mathbb{R}^d$ and output variable $y \in [K] := \{1, \ldots, K\}$, where $K$ is the number of classes.
They follow an unknown joint probability distribution with density $p(\bmx, y)$.
We denote the score function by $\bm g:\mathbb{R}^d \rightarrow \mathbb{R}^K$.
For any test data point $\bmx_0$, our prediction of the output label will be given by $\widehat{y}_0 := \argmax_{z\in[K]} g_z(\bmx_0)$, where $g_z(\cdot)$ is the $z$-th element of $\bm g(\cdot)$, and in case of a tie, $\argmax$ returns the largest argument.
Let $\ell:\mathbb{R}^K \times [K] \rightarrow \mathbb{R}$ denote a loss function.
$\ell$ can be the \emph{zero-one loss},
\begin{equation}
\ell_{01}(\bm v, z') := \begin{cases}
    0 & \text{if}\quad \argmax_{z\in\{1, \dots, K\}} v_z = z',\\
    1 & \text{otherwise},
\end{cases}
\end{equation}
where $\bm v := (v_1, \dots, v_K)^\top \in \mathbb{R}^K$,
or a surrogate loss such as the softmax cross-entropy loss,
\begin{equation}
    \ell_\text{CE}(\bm v, z')
    :=  - \log \frac{\exp(v_{z'})}{\sum_{z\in[K]} \exp(v_z)}.
\end{equation}
For a surrogate loss $\ell$, we denote the \emph{classification risk} by
\begin{equation}
\label{eq:population_risk}
R(\bm g) := \mathbb{E}_{p(\bmx,y)}[\ell(\bm g(\bmx),y)]
\end{equation}
where $\mathbb{E}_{p(\bmx, y)}[\cdot]$ is the expectation over $(\bm x, y) \sim p(\bmx, y)$.
We use $R_{01}(\bm g)$ to denote Eq.~\eqref{eq:population_risk}
when $\ell = \ell_{01}$
and call it the \emph{classification error}.

The goal of multi-class classification is to learn $\bm g$ that minimizes the classification error $R_{01}(\bm g)$.
In optimization, we consider the minimization of the risk with a almost surely differentiable surrogate loss $R(\bm g)$ instead to make the problem more tractable.
Furthermore, since $p(\bmx, y)$ is usually unknown and there is no way to exactly evaluate $R(\bmg)$,
we minimize its empirical version calculated from the training data instead:
\begin{equation}
\label{eq:empirical_risk}
\widehat{R}(\bm g) := \frac{1}{n}\sum^n_{i=1}\ell(\bm g(\bmx_i),y_i),
\end{equation}
where $\{(\bmx_i, y_i)\}_{i=1}^n$ are i.i.d.\@ sampled from $p(\bmx, y)$.
We call $\widehat{R}$ the \emph{empirical risk}.

We would like to clarify some of the undefined terms used in the title and the introduction.
The ``train/test loss'' is the empirical risk with respect to the surrogate loss function $\ell$ over the training/test data, respectively.
We refer to the ``training/test error'' as the empirical risk with respect to $\ell_{01}$ over the training/test data, respectively (which is equal to one minus accuracy) \cite{zhang2004as}.
\footnote{Also see \citet{guo2017icml} for a discussion of the empirical differences of loss and error with neural networks.}

Finally, we formally define the Bayes risk as
\begin{align*}
R^* := \inf_{\bmh} R(\bmh),
\end{align*}
where the infimum is taken over all measurable, vector-valued functions $\bm{h}\colon \mathbb{R}^d \to \mathbb{R}^K$.
The Bayes risk is often referred to as the \emph{Bayes error} if the zero-one loss is used: 
\begin{align*}
\inf_{\bmh} R_{01}(\bmh).
\end{align*}
\subsection{Algorithm}
With flexible models, $\widehat{R}(\bmg)$ w.r.t.\@ a surrogate loss can easily become small if not zero,
as we mentioned in Section~\ref{sec:introduction}; see [C] in Fig.~\ref{fig:overfittings-a}.
We propose a method that ``floods the bottom area and sinks the original empirical risk'' as in Fig.~\ref{fig:overfittings-b} so that the empirical risk cannot go below the flood level.
More technically, if we denote the flood level as $b$, our proposed training objective with flooding is a simple fix.
\begin{definition}

The \emph{flooded empirical risk} is defined as\footnote{
Strictly speaking, Eq.~\eqref{eq:flooding1} is different from Eq.~\eqref{eq:flooding2}, since Eq.~\eqref{eq:flooding1} can be a mini-batch version of Eq.~\eqref{eq:flooding2}.
}
\begin{equation}
\label{eq:flooding2}
\widetilde{R}(\bmg) = |\widehat{R}(\bmg)-b|+b.
\end{equation}
\end{definition}
Note that when $b=0$, then $\widetilde{R}(\bmg) = \widehat{R}(\bmg)$.
The gradient of $\widetilde{R}(\bmg)$ w.r.t.\@ model parameters will point to the same direction as that of $\widehat{R}(\bmg)$ when $\widehat{R}(\bmg) > b$ but in the opposite direction when $\widehat{R}(\bmg) < b$.
This means that when the learning objective is above the flood level, we perform gradient \emph{descent} as usual (gravity zone), but when the learning objective is below the flood level, we perform gradient \emph{ascent} instead (buoyancy zone).

The issue is that in general, we seldom know the optimal flood level in advance.
This issue can be mitigated by searching for the optimal flood level $b^*$ with a hyper-parameter optimization technique.
In practice, we can search for the optimal flood level by performing the exhaustive search in parallel.

\subsection{Implementation}
For large scale problems, we can employ mini-batched stochastic optimization for efficient computation.
Suppose that we have $M$ disjoint mini-batch splits.
We denote the empirical risk~\eqref{eq:empirical_risk} with respect to the $m$-th mini-batch by $\widehat{R}_m(\bmg)$ for $m\in \{1,\ldots, M\}$.
Our mini-batched optimization performs gradient descent updates in the direction of the gradient of 
$\lvert\widehat{R}_\mathrm{m}(\bmg)-b\rvert+b$.
By the convexity of the absolute value function and Jensen's inequality, we have
\begin{equation}
    \widetilde{R}(\bmg) \leq \frac{1}{M}\sum_{m=1}^{M}\left(\lvert\widehat{R}_\mathrm{m}(\bmg)-b\rvert+b\right).
\end{equation}
This indicates that mini-batched optimization will simply minimize an upper bound of the full-batch case with $\widetilde{R}(\bmg)$.

\begin{table*}[t]
\caption{Experimental results for the synthetic data.
The average and standard deviation of the accuracy of each method over $10$ trials.
Sub-table~(A) shows the results without early stopping.
Sub-table~(B) shows the results with early stopping.
The \textbf{boldface} denotes the best and comparable method
in terms of the average accuracy according to the t-test 
at the significance level $1\%$.
The average and standard deviation of the chosen flood level is also shown.
}
\label{tb:synthetics}
\centering
\begin{tabular}{ll|ccc|ccc}
\multicolumn{2}{c}{} & \multicolumn{3}{c}{(A) Without Early Stopping} & \multicolumn{3}{c}{(B) With Early Stopping}\\
\toprule
\multirow{2}{*}{Data} & 
\multirow{2}{*}{\shortstack{Label\\Noise}} & \multirow{2}{*}{\shortstack{Without\\Flooding}} & \multirow{2}{*}{\shortstack{With\\Flooding}} & \multirow{2}{*}{\shortstack{Chosen\\$b$}} & \multirow{2}{*}{\shortstack{Without\\Flooding}} & \multirow{2}{*}{\shortstack{With\\Flooding}} & \multirow{2}{*}{\shortstack{Chosen\\$b$}}\\
& & & & & & &\\
\midrule
   \multirow{3}{*}{\shortstack{Two\\ Gaussians}}  &         Low &           90.52\% (0.71) &  \textbf{92.13\% (0.48)} &  0.17 (0.10) &           90.41\% (0.98) &  \textbf{92.13\% (0.52)} &  0.16 (0.12) \\
   &      Middle &           84.79\% (1.23) &  \textbf{88.03\% (1.00)} &  0.22 (0.09) &           85.85\% (2.07) &  \textbf{88.15\% (0.72)} &  0.23 (0.08) \\
   &        High &           78.44\% (0.92) &  \textbf{83.59\% (0.92)} &  0.32 (0.08) &           81.09\% (2.23) &  \textbf{83.87\% (0.65)} &  0.32 (0.11) \\
\midrule   
     \multirow{3}{*}{Spiral} &         Low &           97.72\% (0.26) &  \textbf{98.72\% (0.20)} &  0.03 (0.01) &  \textbf{97.72\% (0.68)} &  \textbf{98.26\% (0.50)} &  0.02 (0.01) \\
     &      Middle &           89.94\% (0.59) &  \textbf{93.90\% (0.90)} &  0.12 (0.03) &           91.37\% (1.43) &  \textbf{93.51\% (0.84)} &  0.10 (0.04) \\
     &        High &           82.38\% (1.80) &  \textbf{87.64\% (1.60)} &  0.24 (0.05) &           84.48\% (1.52) &  \textbf{88.01\% (0.82)} &  0.22 (0.06) \\
\midrule     
 \multirow{3}{*}{Sinusoid} &         Low &  \textbf{94.62\% (0.89)} &  \textbf{94.66\% (1.35)} &  0.05 (0.04) &  \textbf{94.52\% (0.85)} &  \textbf{95.42\% (1.13)} &  0.03 (0.03) \\
 &      Middle &           87.85\% (1.02) &  \textbf{90.19\% (1.41)} &  0.11 (0.07) &  \textbf{90.56\% (1.46)} &  \textbf{91.16\% (1.38)} &  0.13 (0.07) \\
 &        High &           78.47\% (2.39) &  \textbf{85.78\% (1.34)} &  0.25 (0.08) &  \textbf{83.88\% (1.49)} &  \textbf{85.10\% (1.34)} &  0.22 (0.10) \\
\bottomrule
\end{tabular}
\end{table*}
\begin{table*}[h]
\caption{Benchmark datasets.
Reporting accuracy for all combinations of early stopping and flooding.
We compare ``w/o flood'' and ``w/ flood'' and the better one is shown in \textbf{boldface}.
The best setup for each dataset is shown with \underline{\textbf{underline}}.
``--'' means that flood level of zero was optimal.
``LR'' stands for learning rate and ``aug.'' is an abbreviation of augmentation.
}
\vspace{1mm}
\label{tb:benchmark}
\centering
\begin{tabular}{ll|cc|cc} 
\multicolumn{2}{c}{}&\multicolumn{2}{c}{w/o early stopping} & \multicolumn{2}{c}{w/ early stopping}\\
\toprule
Dataset &Model \& Setup &w/o flood&w/ flood&w/o flood&w/ flood\\
\midrule 
\multirow{3}{*}{MNIST} &MLP& 98.45\% & \textbf{98.76\%} & 98.48\% & \textbf{98.66\%} \\
&MLP w/ weight decay& 98.53\% & \textbf{98.58\%} & 98.51\% & \textbf{98.64\%} \\
&MLP w/ batch normalization& 98.60\% & \underline{\textbf{98.72\%}} & \textbf{98.66\%} & 98.65\% \\ \midrule

\multirow{3}{*}{Kuzushiji}&MLP& 92.27\% & \textbf{93.15\%} & 92.24\% & \textbf{92.90\%} \\
&MLP w/ weight decay& 92.21\% & \textbf{92.53\%} & 92.24\% & \textbf{93.15\%} \\
&MLP w/ batch normalization& 92.98\% & \underline{\textbf{93.80\%}} & 92.81\% & \textbf{93.74\%} \\ \midrule

\multirow{2}{*}{SVHN}&ResNet18 & 92.38\% & \textbf{92.78\%} & 92.41\% & \textbf{92.79\%} \\
&ResNet18 w/ weight decay & 93.20\% & -- & 92.99\% & \underline{\textbf{93.42\%}} \\ \midrule

\multirow{2}{*}{CIFAR-10}
&ResNet44& \textbf{75.38\%} & 75.31\% & 74.98\% & \textbf{75.52\%} \\
&ResNet44 w/ data aug. \& LR decay &88.05\%&\underline{\textbf{89.61\%}}&88.06\%&\textbf{89.48\%}\\
\midrule

\multirow{2}{*}{CIFAR-100}
&ResNet44& \textbf{46.00\%} & 45.83\% & \textbf{46.87\%} & 46.73\% \\
&ResNet44 w/ data aug. \& LR decay & 63.38\% & \underline{\textbf{63.70\%}} & 63.24\% & -- \\
\bottomrule
\end{tabular}
\end{table*}

\begin{figure*}[h]
\centering
\includegraphics[width=\columnwidth*25/40]{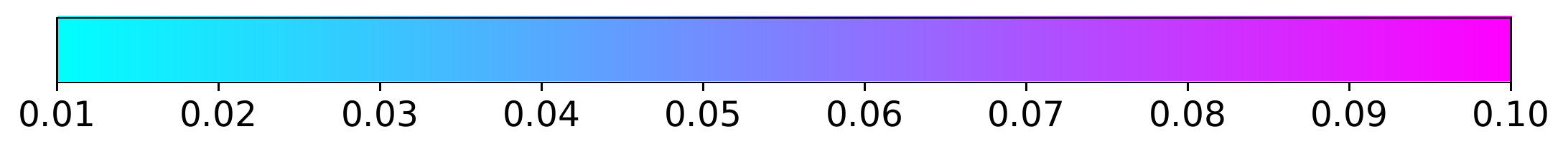}\\
\subcaptionbox{MNIST, MLP}{\includegraphics[width=\columnwidth*20/40]{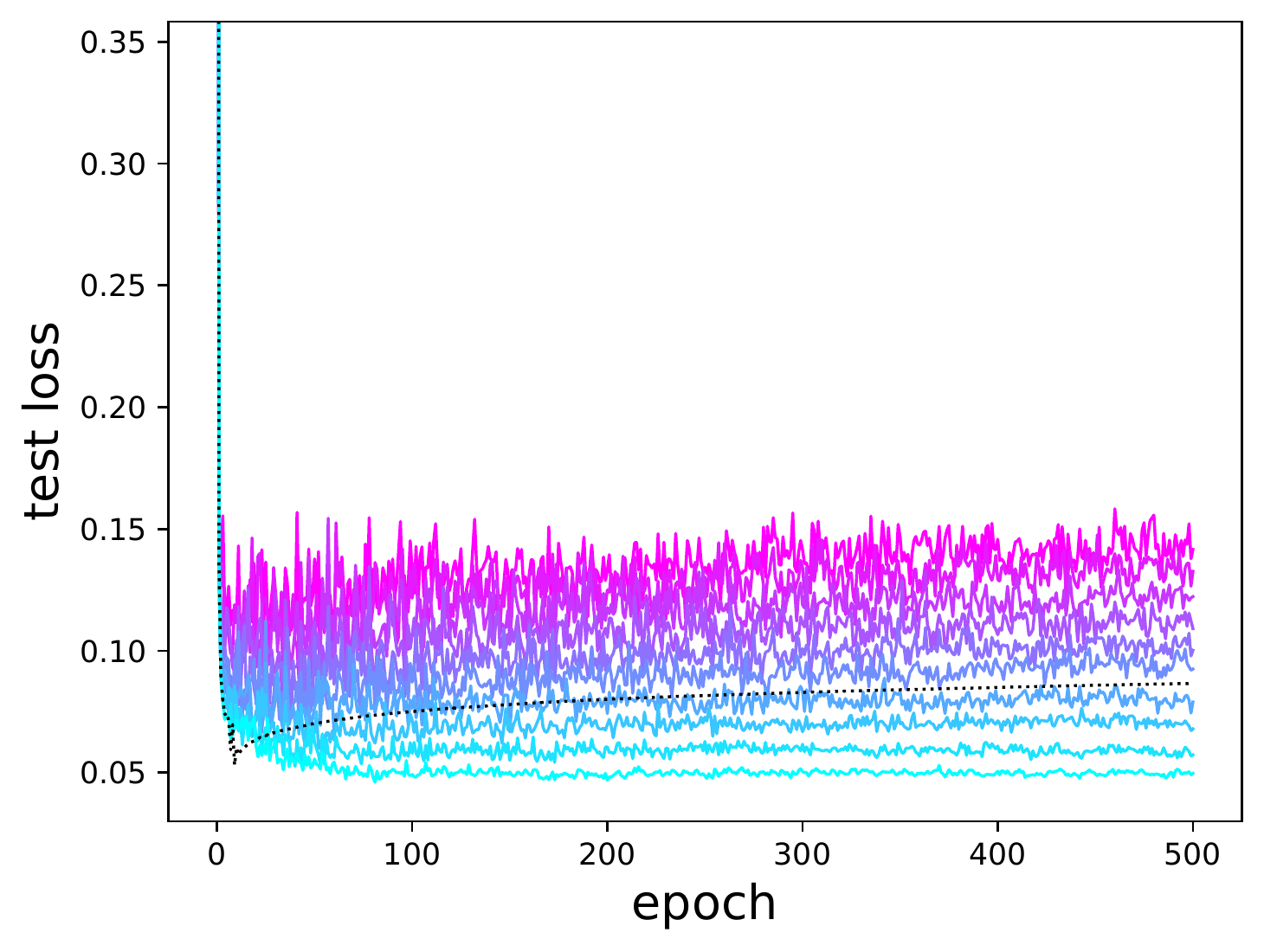}}
\subcaptionbox{KMNIST, MLP}{\includegraphics[width=\columnwidth*20/40]{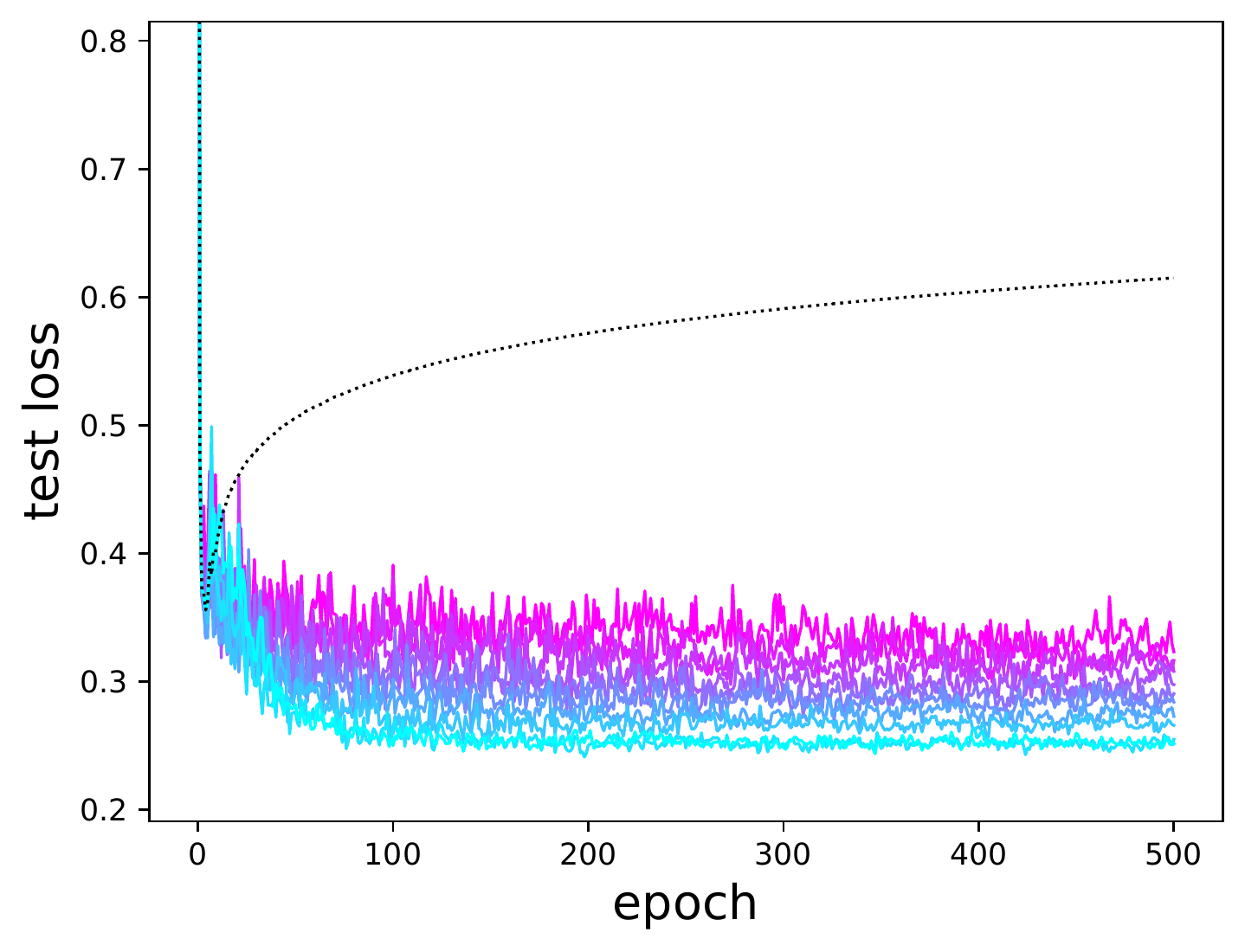}}
\subcaptionbox{SVHN, ResNet18}{\includegraphics[width=\columnwidth*20/40]{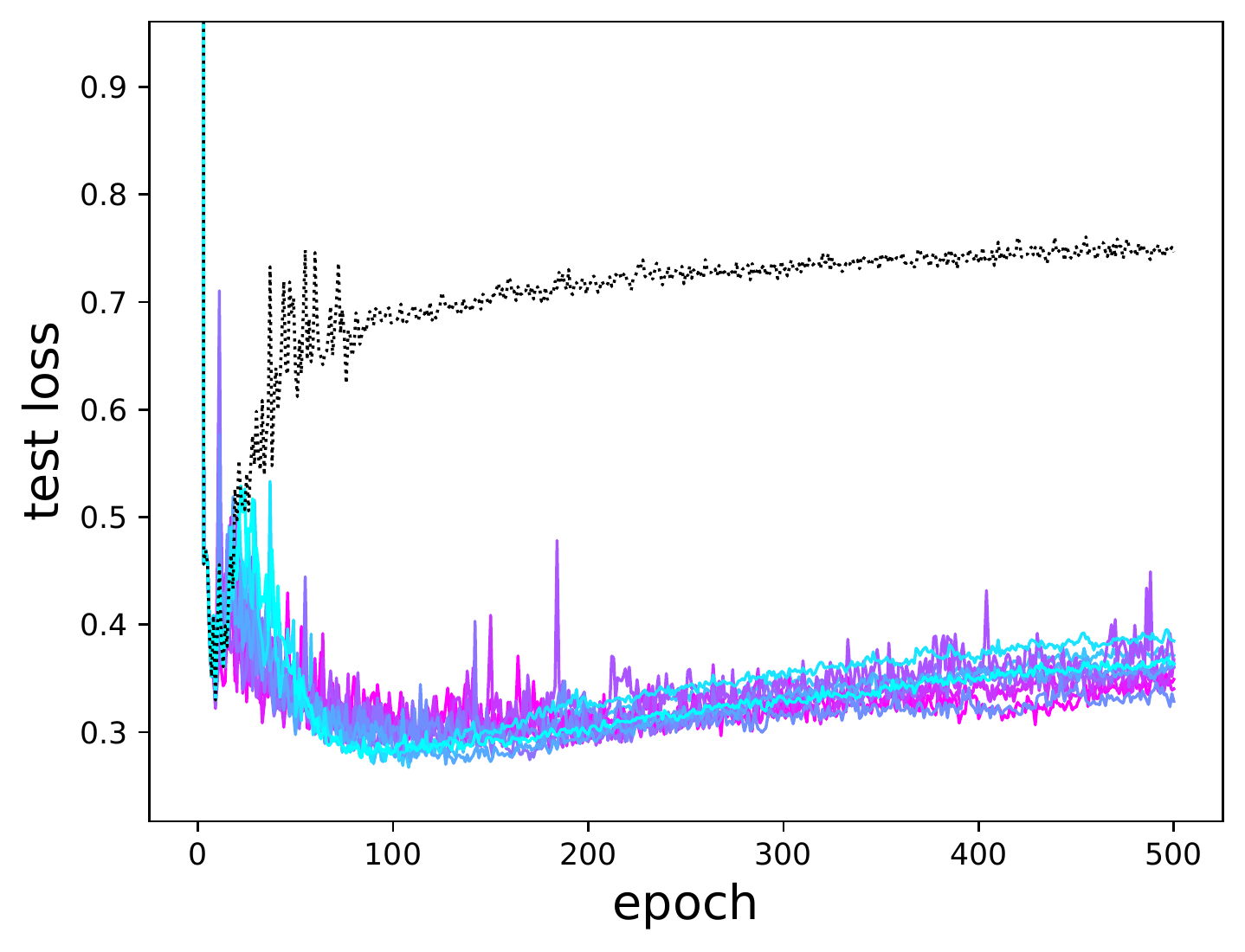}}\\
\subcaptionbox{C10, ResNet44}{\includegraphics[width=\columnwidth*20/40]{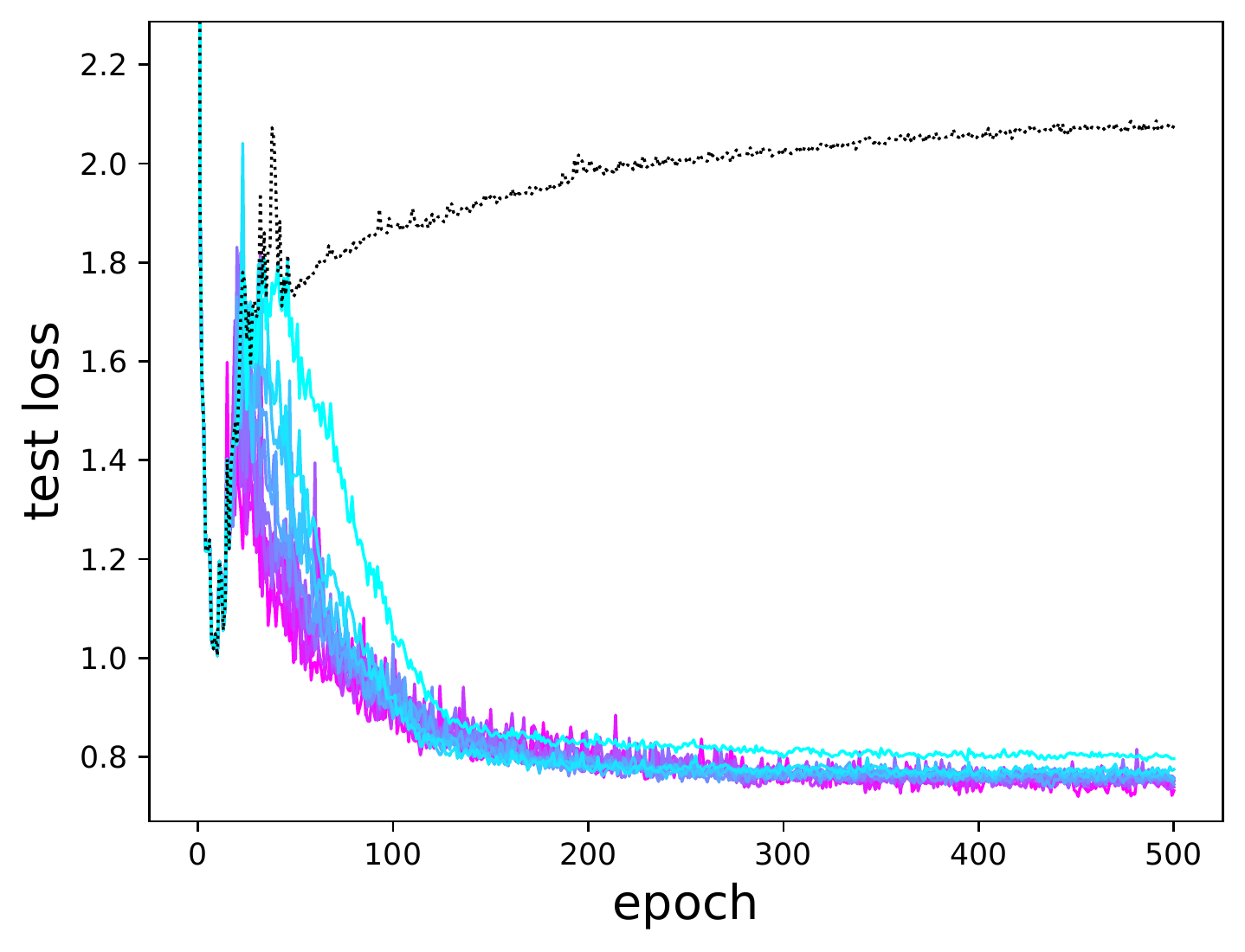}}
\subcaptionbox{C10, ResNet44, DA \& LRD}{\includegraphics[width=\columnwidth*20/40]{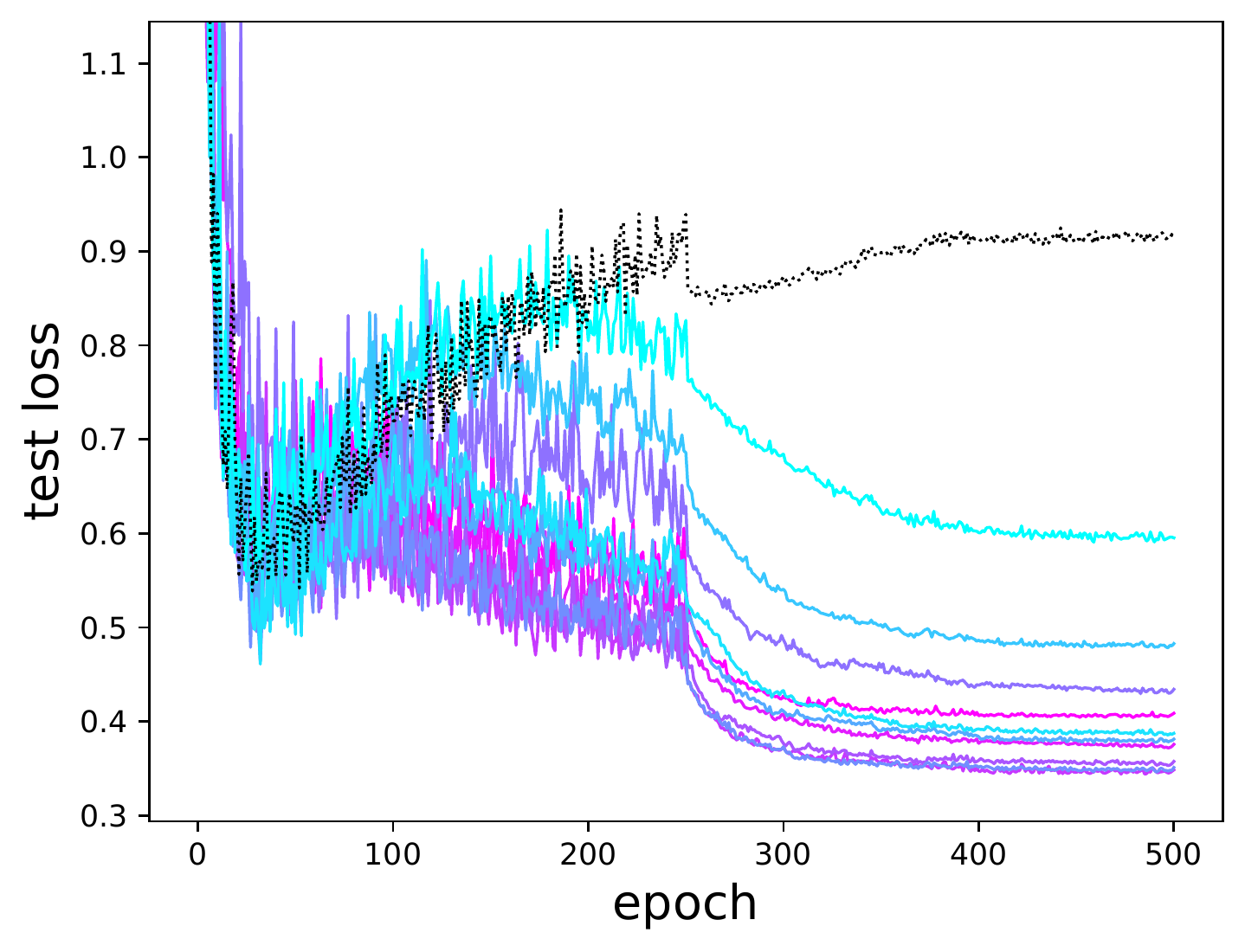}}
\subcaptionbox{C100, ResNet44}{\includegraphics[width=\columnwidth*20/40]{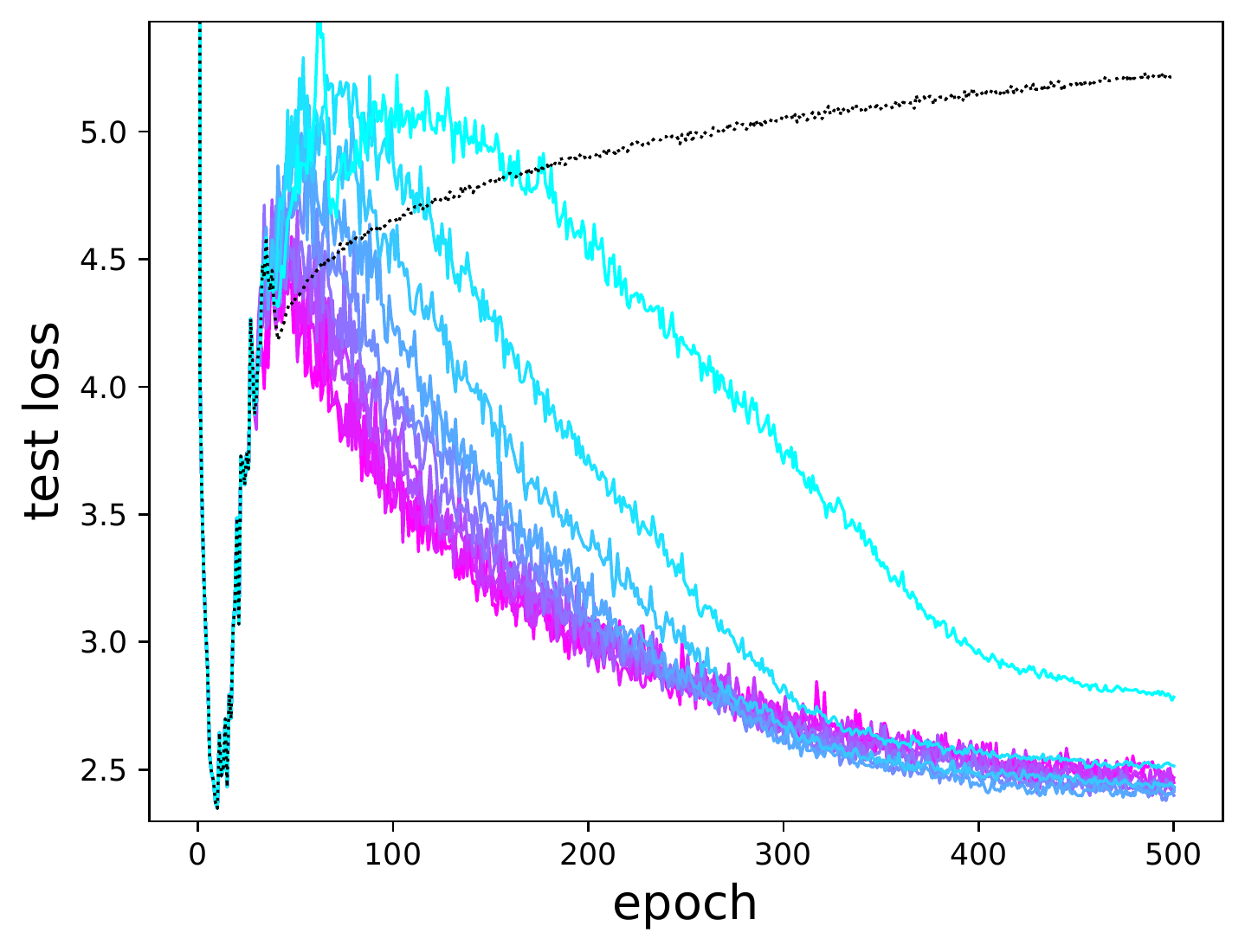}}
\subcaptionbox{C100, ResNet44, DA \& LRD}{\includegraphics[width=\columnwidth*20/40]{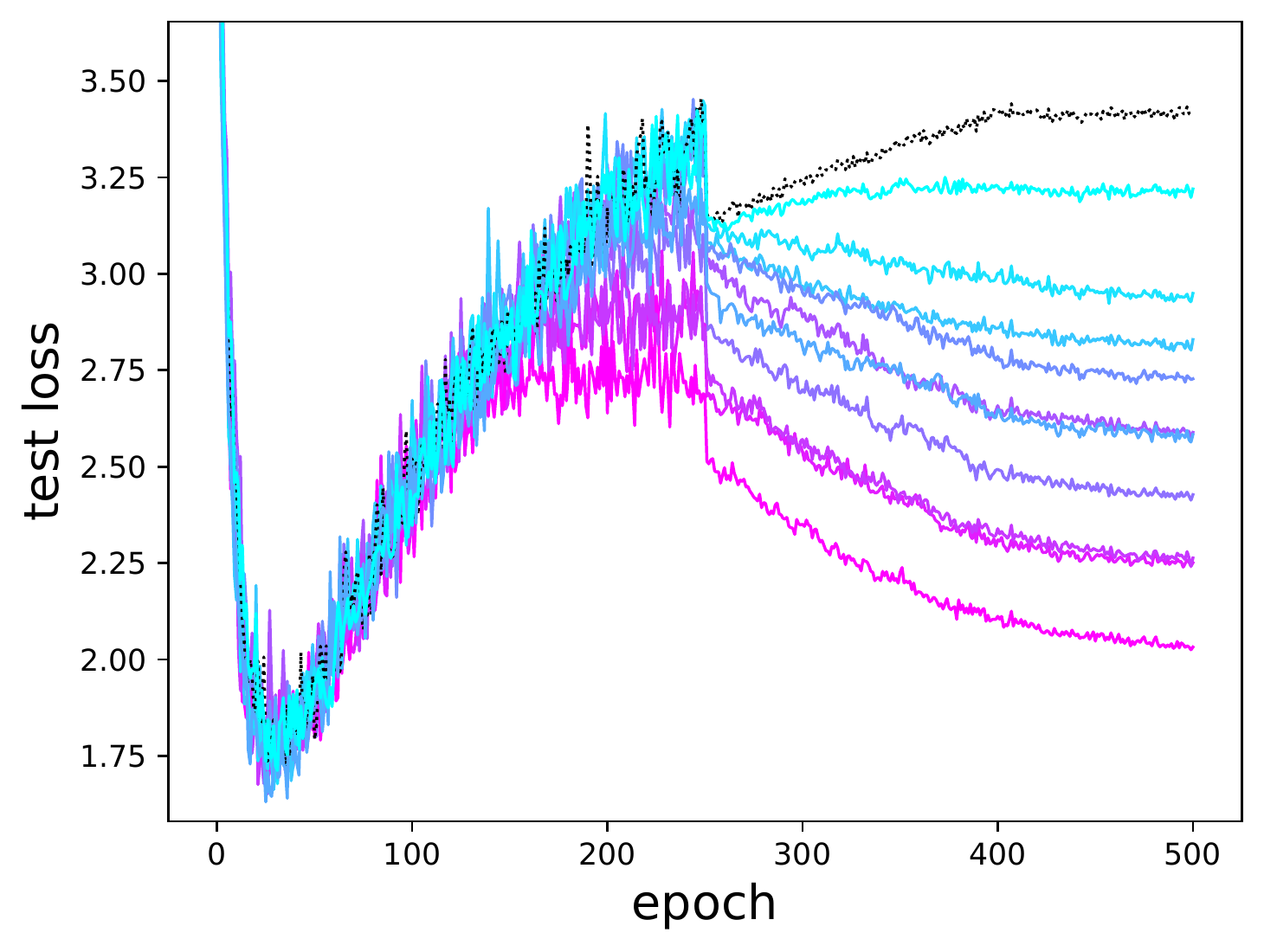}}

\caption{
Learning curves of test loss.
The black dotted line shows the baseline without flooding.
The colored lines show the results for different flooding levels specified by the color bar.
DA and LRD stand for data augmentation and learning rate decay, respectively.
We can observe that adding flooding will lead to lower test loss with a double descent curve in most cases.
See Fig.~\ref{appfig:learning_curves_loss} in Appendix for other datasets.
}
\label{fig:benchmark_learning_curves_0.8_main}
\vskip -0.1in
\end{figure*}
\section{Does Flooding Generalize Better?}
\label{sec:experiments}
In this section, we show experimental results to demonstrate that adding flooding leads to better generalization.
The implementation in this section and the next is based on PyTorch \cite{paszke2019neurips} and demo code is available.\footnote{\url{https://github.com/takashiishida/flooding}}
Experiments were carried out with
NVIDIA GeForce GTX 1080 Ti, NVIDIA Quadro RTX 5000 and Intel Xeon Gold 6142.

\subsection{Synthetic Datasets}\label{sec:synthetic}
The aim of synthetic experiments is to study the behavior of flooding with a controlled setup.

\paragraph{Data}
We use three types of synthetic data:
Two Gaussians, Sinusoid \cite{nakkiran2019neurips}, and Spiral \cite{sugiyama2015introduction}.
Below we explain how these data were generated.

Two Gaussians Data:
We used two $10$-dimensional Gaussian distributions (one for each class)
with covariance matrix identity and means
$\mu_{\mathrm{P}}=[0,0,\ldots,0]^\top$ and $\mu_{\mathrm{N}}=[m,m,\ldots,m]^\top$,
where $m=1.0$.
The training, validation, and test sample sizes are $100$, $100$, and $20000$, respectively.

Sinusoid Data:
The sinusoid data \cite{nakkiran2019neurips} are generated as follows.
We first draw input data points from the $2$-dimensional standard Gaussian distribution,
i.e., $\bm{x} \sim N(\bm{0}_2,\bm{I}_2)$, where $\bm{0}_2$ is the two-dimensional zero vector, and $\bm{I}_2$ $2\times2$ identity matrix.
Then put class labels based on  
\begin{align*}  
    y = \mathrm{sign}(\bm{x}^\top\bm{w} + \sin(\bm{x}^\top\bm{w'})) ,
\end{align*}
where $\bm{w}$ and $\bm{w'}$ are any two $2$-dimesional vectors such that $\bm{w}\perp\bm{w'}$.
The training, validation, and test sample sizes are $100$, $100$, and $20000$, respectively.

Spiral Data:
The spiral data~\citep{sugiyama2015introduction} is two-dimensional synthetic data.
Let $\theta^+_1 := 0, \theta^+_2, \dots, \theta^+_{n^+} := 4\pi$ be equally spaced $n^+$ points in the interval $[0, 4\pi]$,
and $\theta^-_1 := 0, \theta^-_2, \dots, \theta^-_{n^-} := 4\pi$ be equally spaced $n^-$ points in the interval $[0, 4\pi]$.
Let positive and negative input data points be
\begin{align*}
    \bm{x}^+_i &:= \theta_i [\cos(\theta_i), \sin(\theta_i)]^\top + \tau \bm{\nu}^+_i,\\
    \bm{x}^-_i &:= (\theta + \pi) [\cos(\theta), \sin(\theta)]^\top + \tau \bm{\nu}^-_i
\end{align*}
for $i = 1, \dots, n$, where $\tau$ controls the magnitude of the noise, $\bm{\nu}^+_i$ and $\bm{\nu}^-_i$ are i.i.d.\@ distributed according to the two-dimensional standard normal distribution.
Then, we make data for classification by
$\{(\bm x_i, y_i)\}_{i=1}^n := \{(\bm x^+_i, +1)\}_{i^+=1}^{n^+}\cup\{(\bm x^-_i, -1)\}_{i^-=1}^{n^-}$,
where $n := n^+ + n^-$.
The training, validation, and test sample sizes are $100$, $100$, and $10000$ per class respectively.

\paragraph{Settings}
We use a five-hidden-layer feedforward neural network with $500$ units in each hidden layer with the ReLU activation function \citep{nair10icml}.
We train the network for $500$ epochs with the logistic loss and the Adam \citep{kingma15iclr} optimizer with $100$ mini-batch size and learning rate of $0.001$.
The flood level is chosen from $b\in\{ 0, 0.01, 0.01, \ldots, 0.50 \}$.
We tried adding label noise to dataset, by flipping $1\%$ (Low), $5\%$ (Middle), or $10\%$ (High) of the labels randomly.
This label noise corresponds to varying the Bayes risk, i.e., the difficulty of classification.
Note that the training with $b=0$ is identical to the baseline method without flooding.
We report the test accuracy of the flood level with the best validation accuracy.
We first conduct experiments without early stopping, which means that the last epoch was chosen for all flood levels.

\paragraph{Results}
The average and standard deviation of the accuracy of each method over $10$ trials are summarized in Table~\ref{tb:synthetics}.
It is worth noting that the average of the chosen flood level $b$ is larger for the larger label noise because the increase of label noise is expected to increase the Bayes risk.
This implies the positive correlation between the optimal flood level and the Bayes risk.

From (A) in Table~\ref{tb:synthetics}, we can see that the method with flooding often improves test accuracy over the baseline method without flooding.
As we mentioned in the introduction, it can be harmful to keep training a model until the end without flooding.
However, with flooding, the model at the final epoch has good prediction performance according to the results,
which implies that flooding helps the late-stage training improve test accuracy.

We also conducted experiments with early stopping, meaning that we chose the model that recorded the best validation accuracy during training.
The results are reported in sub-table (B) of Table~\ref{tb:synthetics}.
Compared with sub-table (A), we see that early stopping improves the baseline method without flooding in many cases.
This indicates that training longer without flooding was harmful in our experiments.
On the other hand, the accuracy for flooding combined with early stopping is often close to that with early stopping,
meaning that training until the end with flooding tends to be already as good as doing so with early stopping. The table shows that flooding often improves or retains the test accuracy of the baseline method without flooding even after deploying early stopping.
Flooding does not hurt performance but can be beneficial for methods used with early stopping.

\subsection{Benchmark Experiments}
\label{sec:benchmark}
We next perform experiments with benchmark datasets.
Not only do we compare with the baseline without flooding, we also compare or combine with other general regularization methods, which are early stopping and weight decay.

\paragraph{Settings}
We use the following benchmark datasets: MNIST, Kuzushiji-MNIST, SVHN, CIFAR-10, and CIFAR-100.
The details of the benchmark datasets can be found in Appendix~\ref{appsec:bench_datasets}.
Stochastic gradient descent~\cite{robbins1951} is used with learning rate of 0.1 and momentum of 0.9 for 500 epochs.
For MNIST and Kuzushiji-MNIST, we use a two-hidden-layer feedforward neural network with 1000 units and the ReLU activation function \cite{nair10icml}.
We also compared adding batch normalization \citep{ioffe2015icml}.
For SVHN, we used ResNet-18 from \citet{he2016cvpr} with the implementation provided in \citet{paszke2019neurips}.
For MNIST, Kuzushiji-MNIST, and SVHN, we compared adding weight decay with rate $1\times 10^{-5}$.
For CIFAR-10 and CIFAR-100, we used ResNet-44 from \citet{he2016cvpr} with the implementation provided in \citet{akamasterrepo}.
For CIFAR-10 and CIFAR-100, we compared adding simple data augmentation (random crop and horizontal flip) and learning rate decay (multiply by 0.1 after 250 and 400 epochs).
We split the original training dataset into training and validation data with with a proportion of $80:20$ except for when we used data augmentation, we used $85:15$.
We perform the exhaustive hyper-parameter search for the flood level with candidates from
$\{0.00, 0.01, \ldots, 0.10\}$.
We used the original labels and did not add label noise.
We deployed early stopping in the same way as in Section~\ref{sec:synthetic}.

\paragraph{Results}
We show the results in Table~\ref{tb:benchmark}. For each dataset, the best performing setup and combination of regularizers always use flooding.
Combining flooding with early stopping, weight decay, data augmentation, batch normalization, and/or learning rate decay usually has complementary effects.
As a byproduct, we were able to produce a double descent curve for the test loss with a relatively few number of epochs, shown in Fig.~\ref{fig:benchmark_learning_curves_0.8_main}.

\section{\emph{Why} Does Flooding Generalize Better?}
In this section, we investigate four key properties of flooding to seek potential reasons for better generalization.
\subsection{Memorization}
\begin{figure}[ht]
\centering
\subcaptionbox{w/o early stopping}{\includegraphics[width=\columnwidth*19/40]{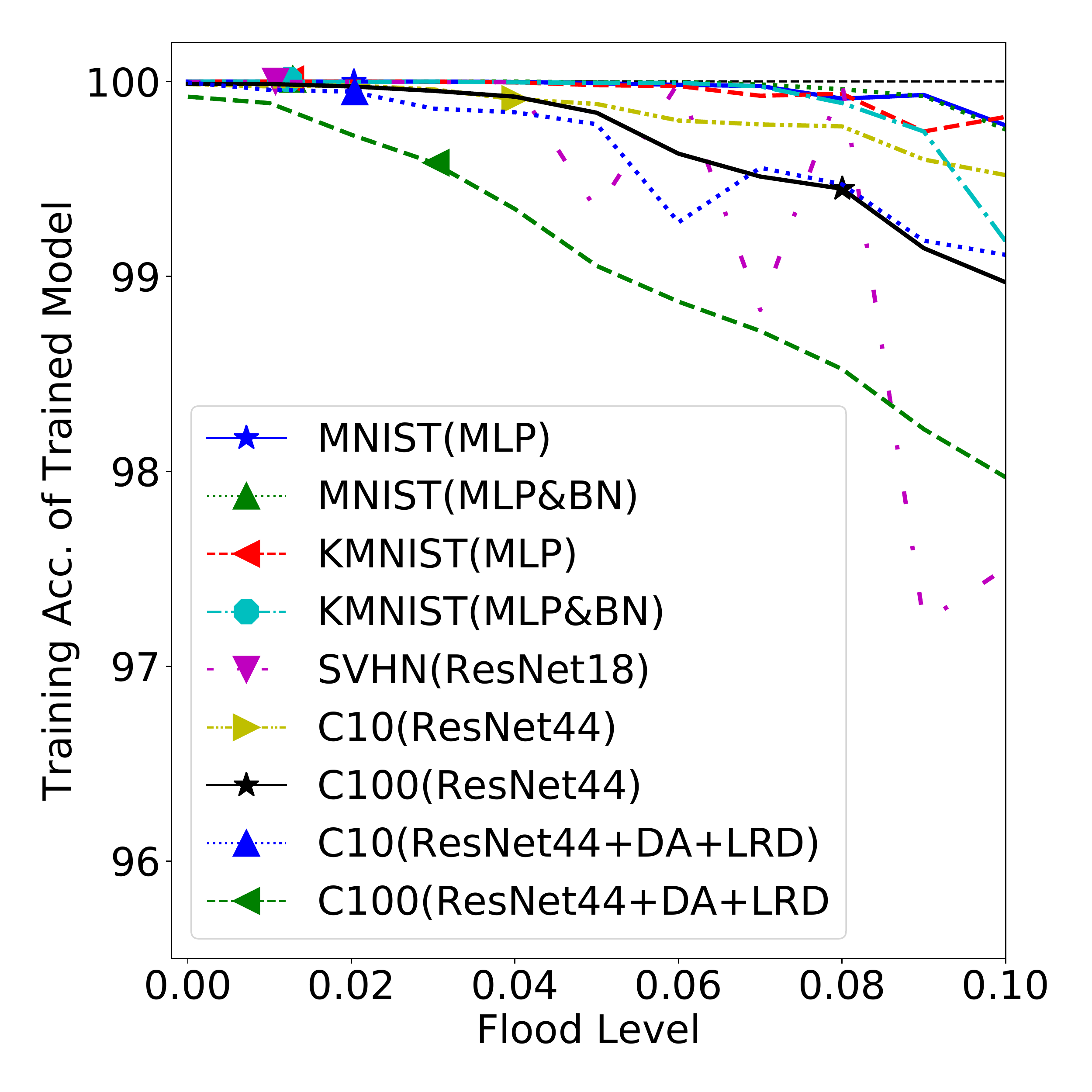}}
\subcaptionbox{w/ early stopping}{\includegraphics[width=\columnwidth*19/40]{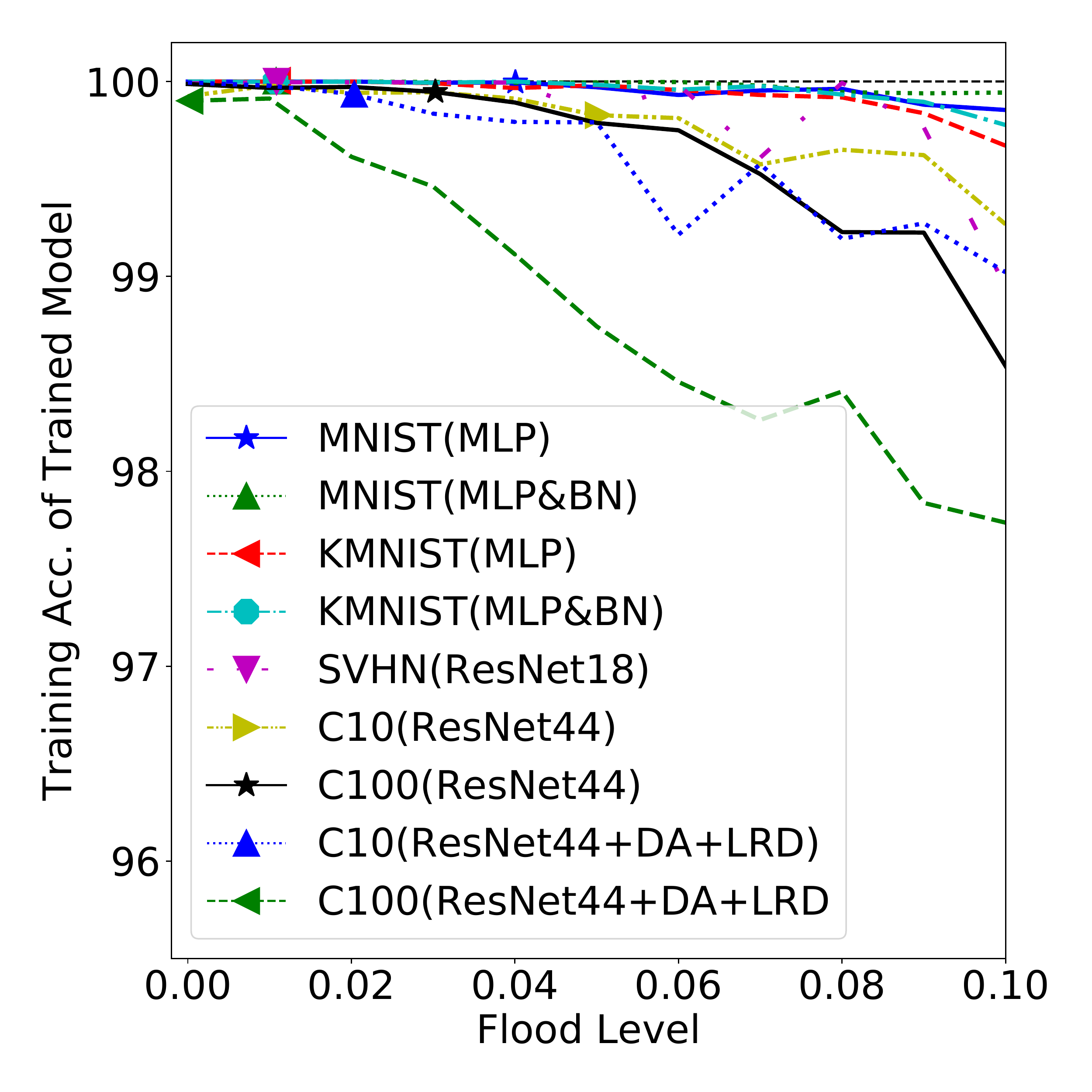}}
\caption{
Vertical axis is the training accuracy and the horizontal axis is the flood level.
Marks are placed on the flood level that was chosen based on validation accuracy.
}
\label{fig:memorization}
\end{figure}

Can we maintain memorization \cite{zhang2017iclr,arpit2017icml,belkin2018neurips} even after adding flooding?
We investigate if the trained model has zero training error (100\% accuracy) for the flood level that was chosen with validation data.
The results are shown in Fig.~\ref{fig:memorization}.

All datasets and settings show downward curves, implying that the model will give up eventually on memorizing all training data as the flood level becomes higher.
A more interesting observation is the position of the optimal flood level (the one chosen by validation accuracy which is marked with $\star$, $\triangle$, $\triangleleft$, $\circ$, $\triangledown$ or $\triangleright$).
We can observe that the marks are often plotted at zero error.
These results are consistent with recent empirical works that imply zero training error leads to lower generalization error \cite{belkin2019pnas, nakkiran2020iclr},
but we further demonstrate that zero training loss may be harmful under zero training error.

\subsection{Performance and gradients}
\label{subsec:gradients}
\begin{figure*}[ht]
\centering
\subcaptionbox{MNIST, x:train}{\includegraphics[width=\columnwidth*39/80]{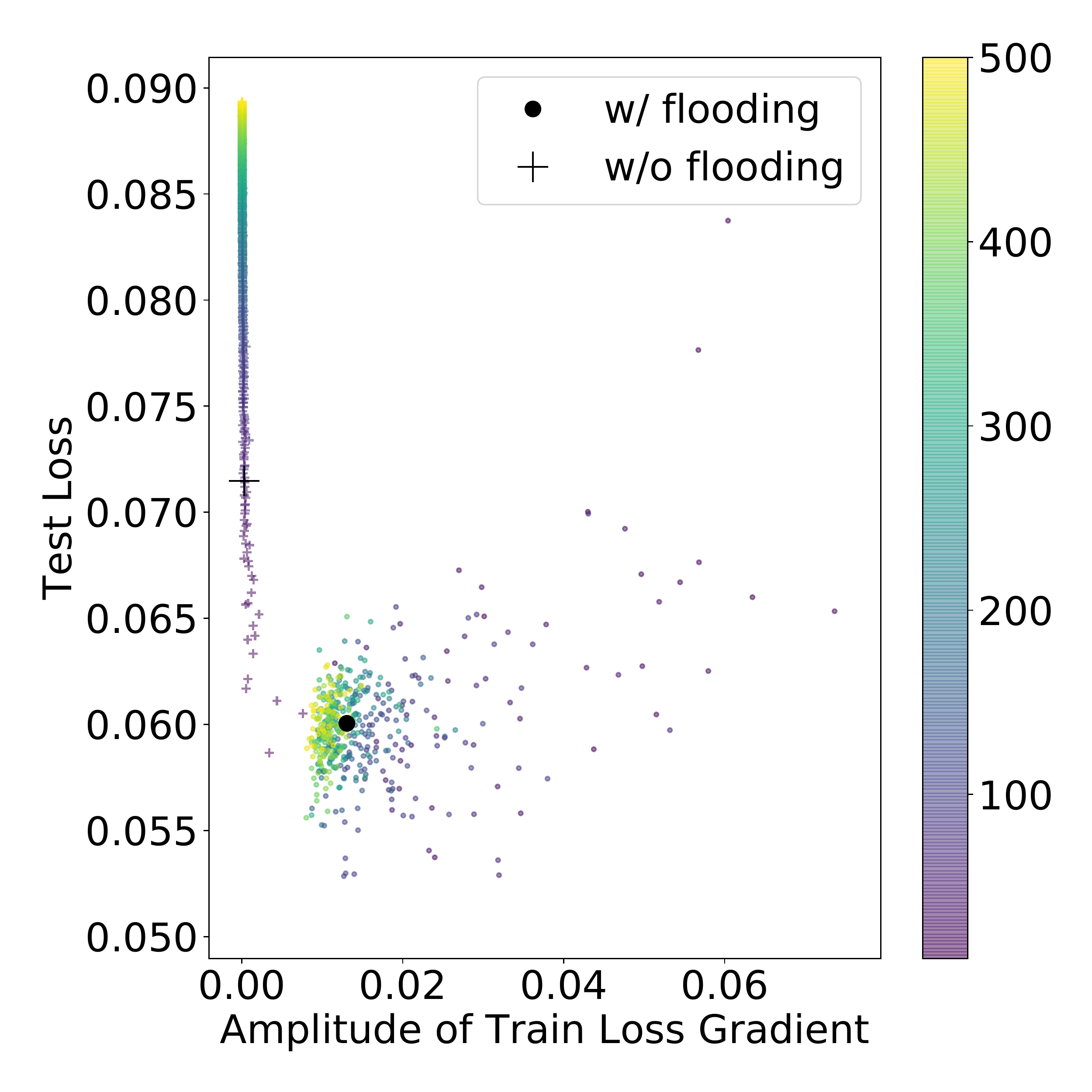}}
\subcaptionbox{MNIST, x:test}{\includegraphics[width=\columnwidth*39/80]{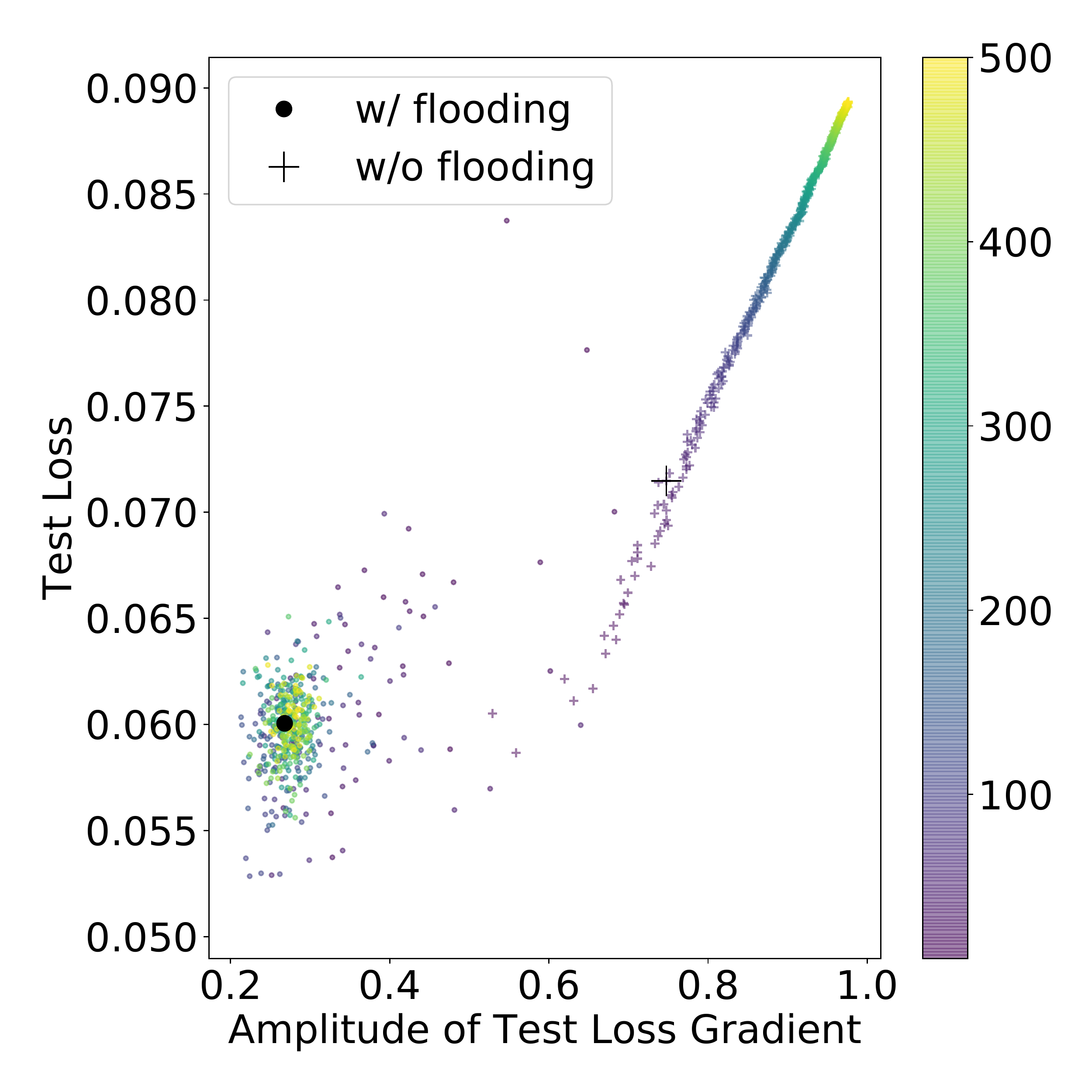}}
\subcaptionbox{CIFAR-10, x:train}{\includegraphics[width=\columnwidth*39/80]{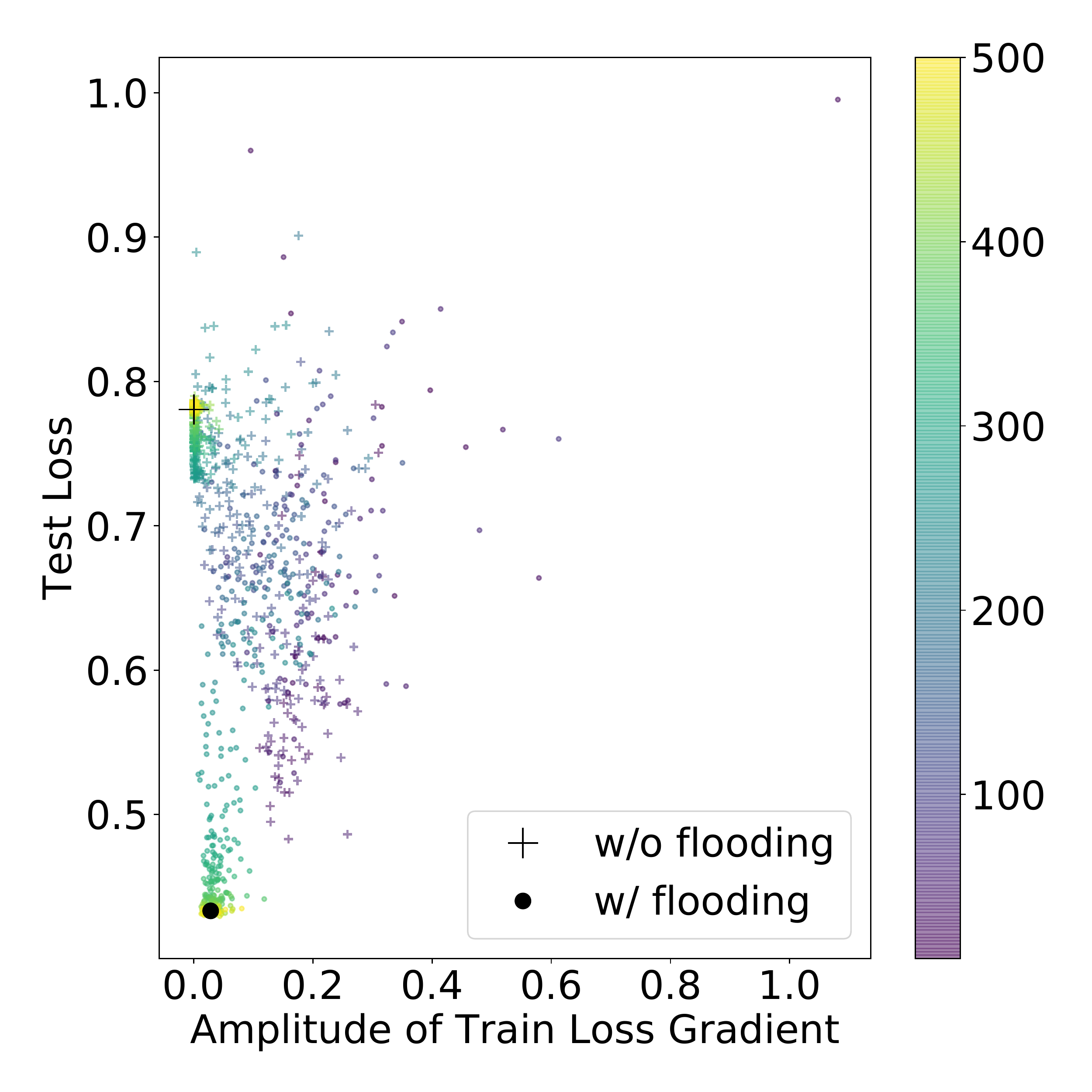}}
\subcaptionbox{CIFAR-10, x:test}{\includegraphics[width=\columnwidth*39/80]{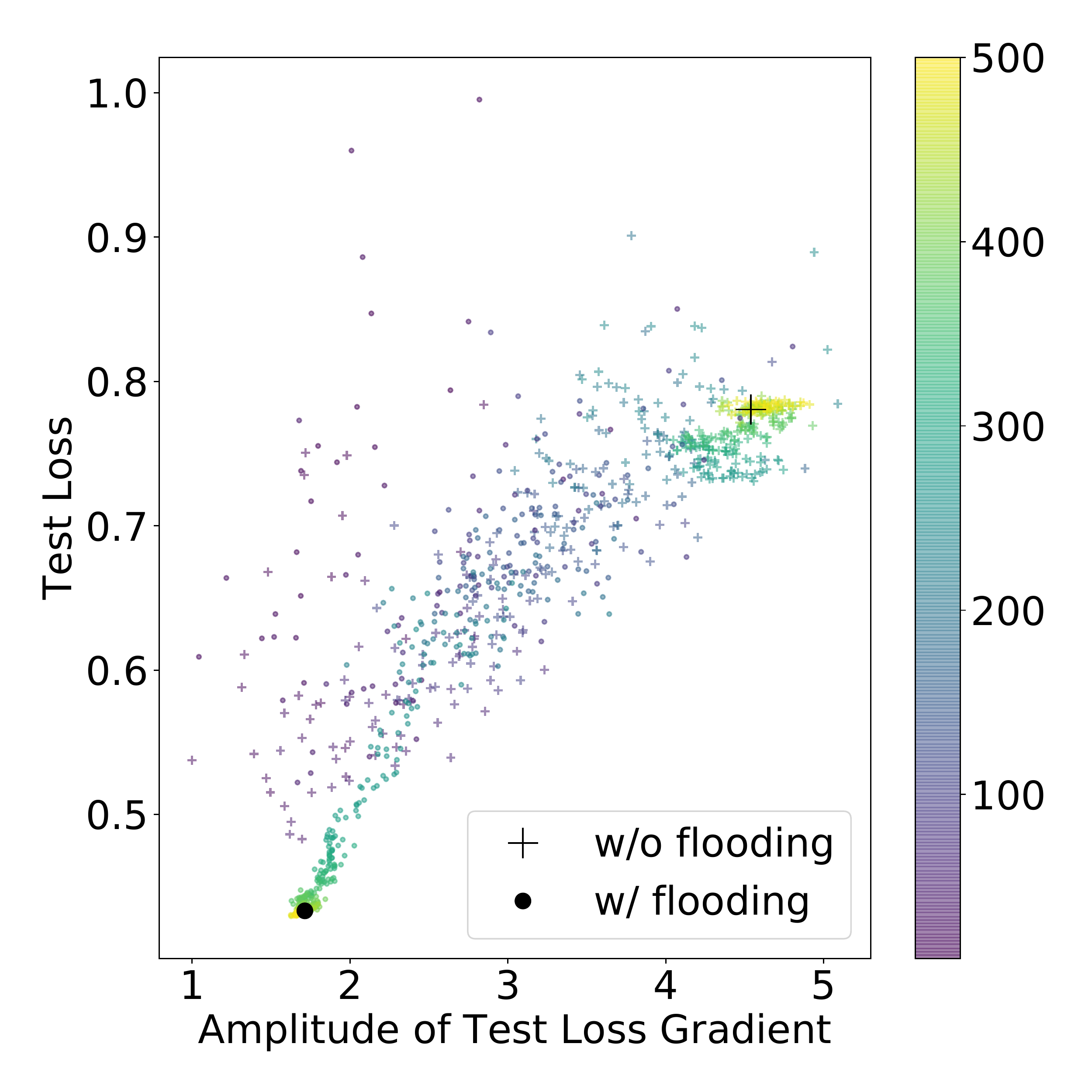}}
\caption{
Relationship between test loss and amplitude of gradient (with training or test loss).
Each point (``$\circ$'' or ``$+$'') in the figures corresponds to a single model at a certain epoch.
We remove the first 10 epochs and plot the rest.
``$\circ$'' is used for the method with flooding  and ``$+$'' is used for the method without flooding.
The large black ``$\circ$'' and ``$+$'' show the epochs with early stopping.
The color becomes lighter (purple $\rightarrow$ yellow) as the training proceeds.
See Fig.~\ref{appfig:performance_gradients} in Appendix for other datasets.
}
\label{fig:performance_gradients_main}
\end{figure*}

We visualize the relationship between test performance (loss or accuracy) and gradient amplitude of the training/test loss in Fig.~\ref{fig:performance_gradients_main}, where the gradient amplitude is the $\ell_2$ norm of the \emph{filter-normalized gradient} of the loss.
The filter-normalized gradient is the gradient appropriately scaled depending on the magnitude of the corresponding convolutional filter, similarly to \cite{li2018neurips}.
More specifically, for each filter of every convolutional layer, we multiply the corresponding elements of the gradient by the norm of the filter.
Note that a fully connected layer is a special case of convolutional layer and is also subject to this scaling.

For the sub-figures with gradient amplitude of training loss on the horizontal axis, ``$\circ$'' marks (w/ flooding) are often plotted on the right of ``$+$'' marks (w/o flooding), which implies that flooding prevents the model from staying at a local minimum.
For the sub-figures with gradient amplitude of test loss on the horizontal axis, the method with flooding (``$\circ$'') improves performance while the gradient amplitude becomes smaller.

\subsection{Flatness}
\label{subsec:flatness}
We give a one-dimensional visualization of flatness \cite{li2018neurips}.
We compare the flatness of the model right after the empirical risk with respect to a mini-batch becomes smaller than the flood level, $\widehat{R}_m(\bmg)<b$, for the first time (dotted blue) and the model after training (solid blue).
We also compare them with the model trained by the baseline method without flooding, and training is finished (solid red).
\begin{figure*}[ht]
\subcaptionbox{SVHN (train)}{\includegraphics[width=\columnwidth*39/80]{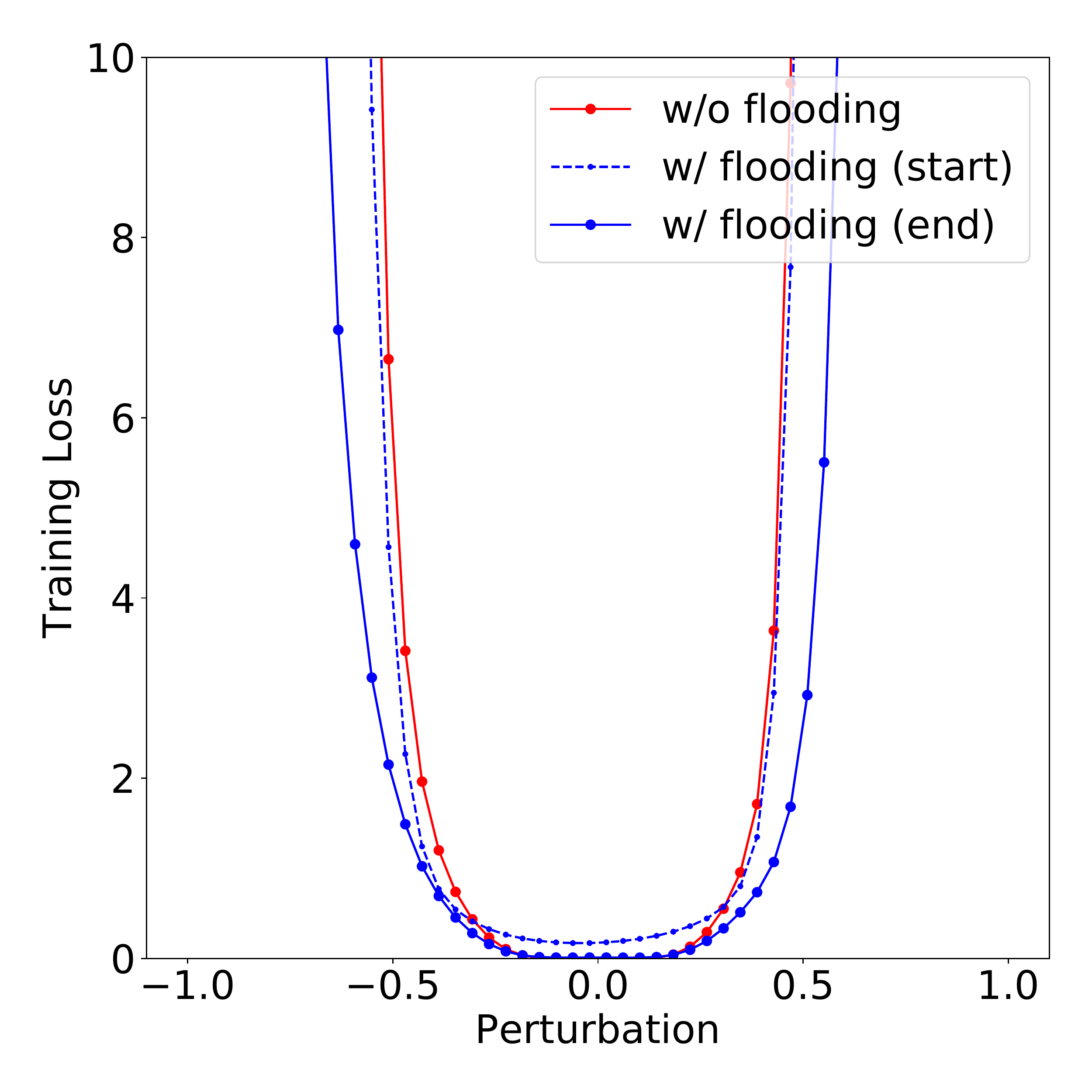}}
\subcaptionbox{SVHN (test)}{\includegraphics[width=\columnwidth*39/80]{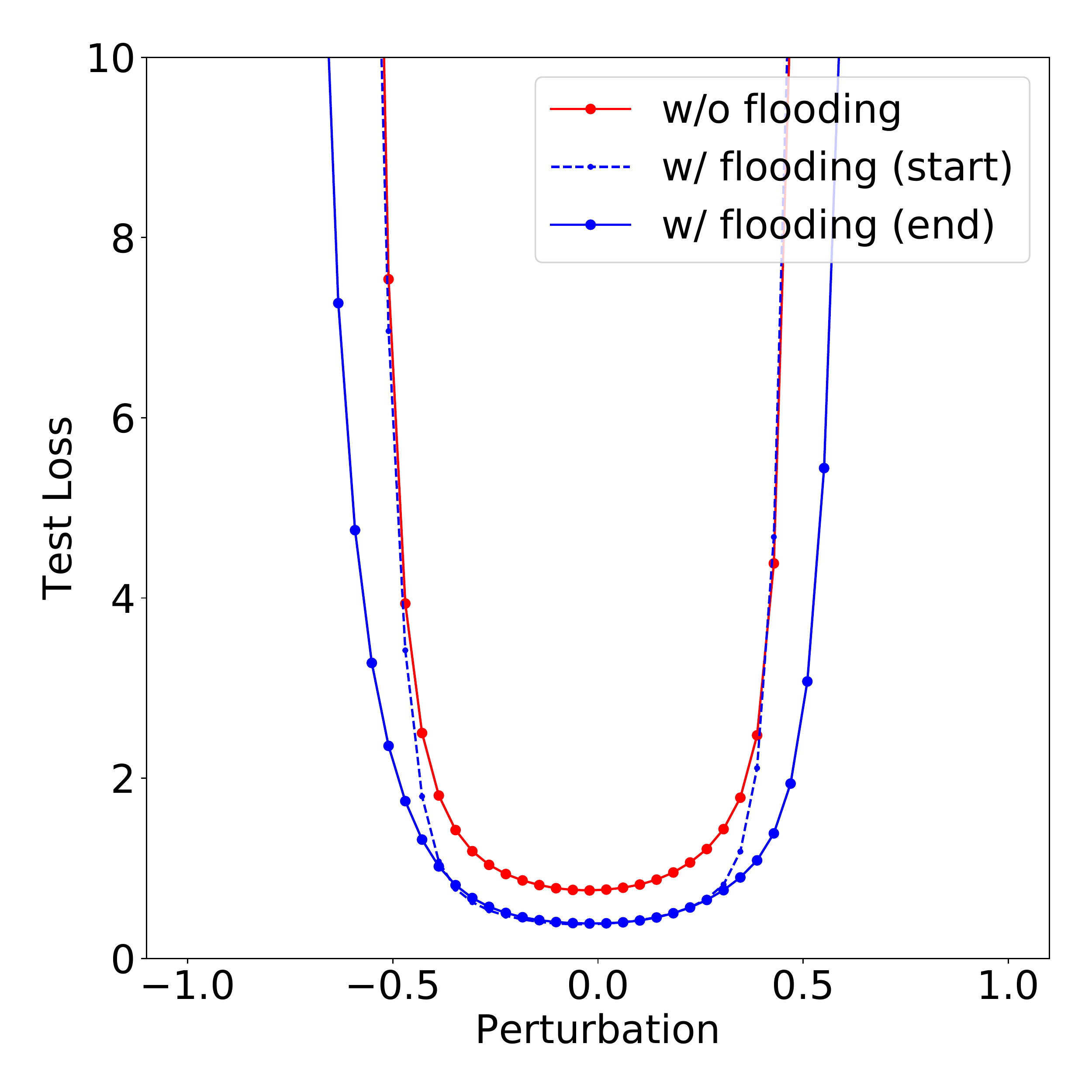}}
\subcaptionbox{CIFAR-10 (train)}{\includegraphics[width=\columnwidth*39/80]{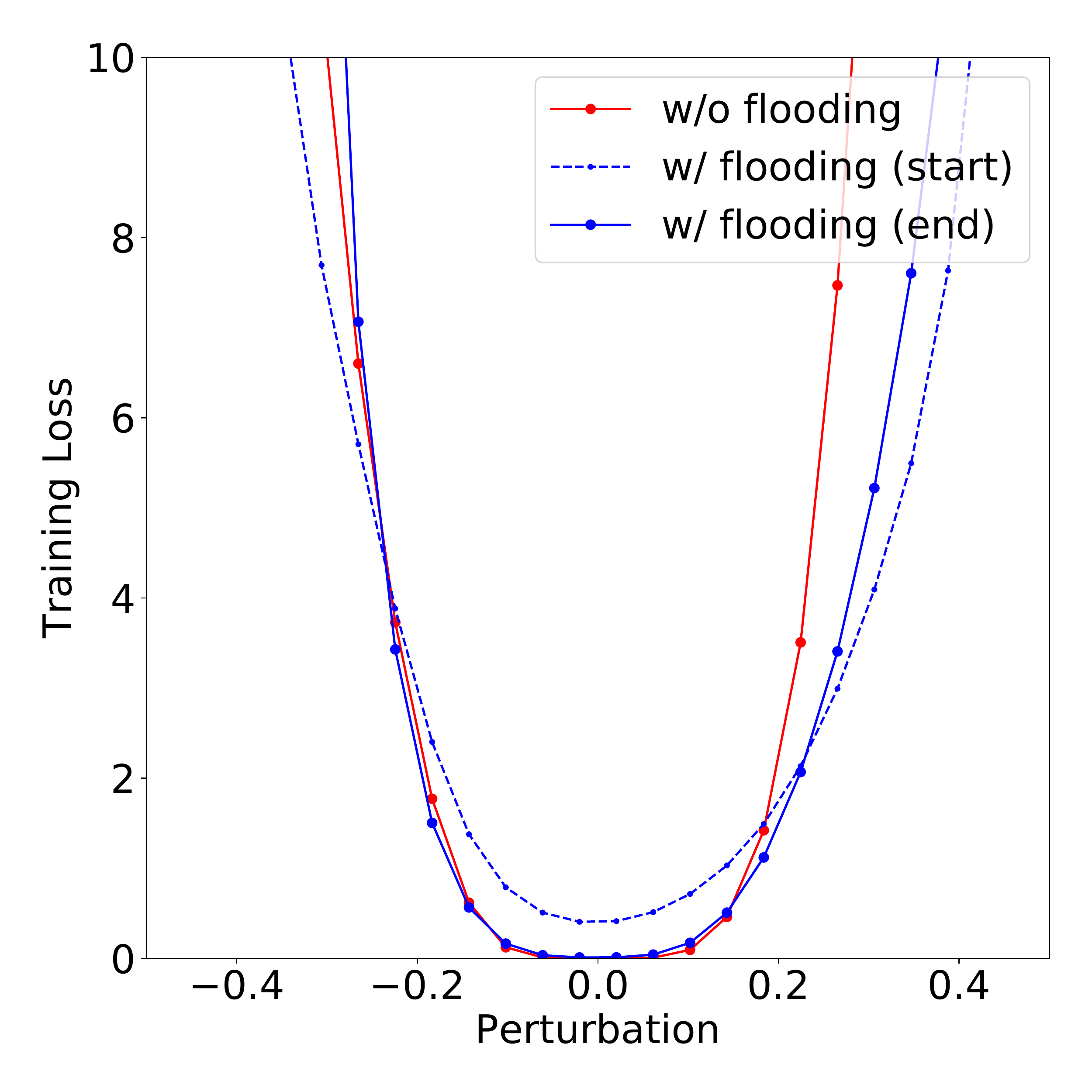}}
\subcaptionbox{CIFAR-10 (test)}{\includegraphics[width=\columnwidth*39/80]{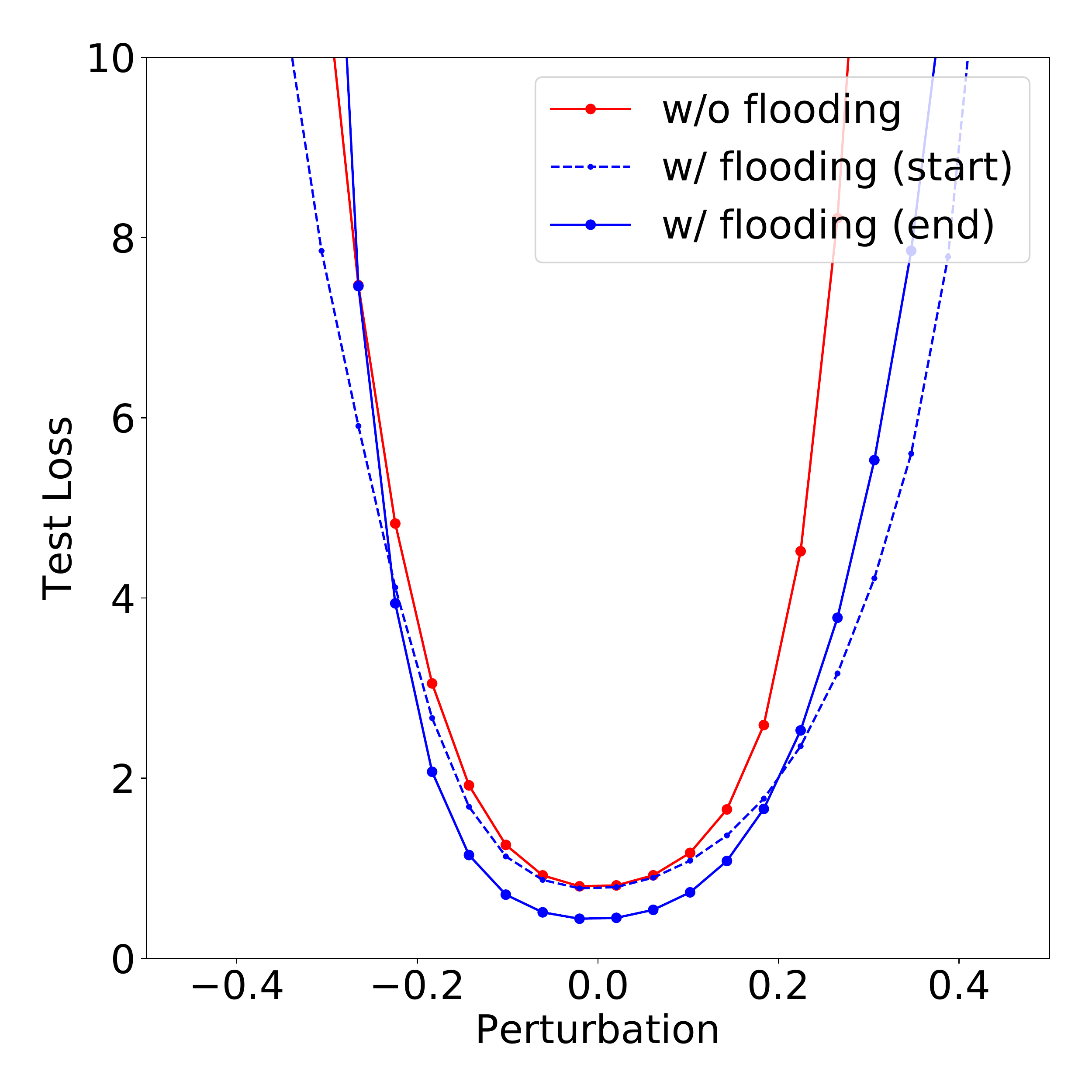}}
\caption{
One dimensional visualization of flatness of training/test loss with respect to perturbation.
We depict the results for 3 models: the model when the empirical risk with respect to training data is below the flooding level for the first time during training(dotted blue), the model at the end of training with flooding (solid blue), and the model at the end of training without flooding (solid red).
See Fig.~\ref{appfig:flatness} in Appendix for other datasets.
}
\label{fig:flatness_main}
\end{figure*}

In Fig.~\ref{fig:flatness_main}, the test loss becomes lower and flatter during the training with flooding.
Note that the training loss, on the other hand, continues to float around the flood level until the end of training after it enters the flooding zone.
We expect that the model makes a random walk and escapes regions with sharp loss landscapes during the period.
This may be a possible reason for better generalization results with our proposed method.

\subsection{Theoretical Insight}
We show that the mean squared error (MSE) of the flooded risk estimator is smaller than that of the original risk estimator without flooding, with the condition that the flood level is between the original training loss and the test loss, in the following theorem.
\begin{theorem}\label{theorem:mse}
Fix any measurable vector-valued function $\bmg$.
If the flood level $b$ satisfies $b \le R(\bmg)$, we have
\begin{equation}
    \mathrm{MSE}(\widehat{R}(\bmg)) \ge \mathrm{MSE}(\widetilde{R}(\bmg)).
\end{equation}
If $b$ further satisfies $b < R(\bmg)$ and $\Pr[\widehat{R}(\bmg) < b] >0$, the inequality will be strict:
\begin{equation}
    \mathrm{MSE}(\widehat{R}(\bmg)) > \mathrm{MSE}(\widetilde{R}(\bmg)).
\end{equation}
\end{theorem}

See Appendix~\ref{appsec:mse_proof} for formal discussions and the proof.

\section{Conclusion}
We proposed a novel regularization method called \emph{flooding} that keeps the training loss to stay around a small constant value, to avoid zero training loss.
In our experiments, the optimal flood level often maintained memorization of training data, with zero error.
With flooding, our experiments confirmed that the test accuracy improves for various synthetic and benchmark datasets, and we theoretically showed that the MSE will be reduced under certain conditions.

As a byproduct, we were able to produce a double descent curve for the test loss when flooding was used.
An important future direction is to study the relationship between this and the double descent curves from previous works \cite{krogh1992neurips,belkin2019pnas,nakkiran2020iclr}.

\section*{Acknowledgements}
We thank Chang Xu, Genki Yamamoto, Jeonghyun Song, Kento Nozawa, Nontawat Charoenphakdee, Voot Tangkaratt, and Yoshihiro Nagano for the helpful discussions.
We also thank anonymous reviewers for helpful comments.
TI was supported by the Google PhD Fellowship Program and JSPS KAKENHI 20J11937.
MS and IY were supported by JST CREST Grant Number JPMJCR18A2 including AIP challenge program, Japan.

\bibliography{example_paper}

\begin{thebibliography}{58}
\providecommand{\natexlab}[1]{#1}
\providecommand{\url}[1]{\texttt{#1}}
\expandafter\ifx\csname urlstyle\endcsname\relax
  \providecommand{\doi}[1]{doi: #1}\else
  \providecommand{\doi}{doi: \begingroup \urlstyle{rm}\Url}\fi

\bibitem[Arpit et~al.(2017)Arpit, Jastrzebski, Ballas, Krueger, Bengio, Kanwal,
  Maharaj, Fischer, Courville, Bengio, and Lacoste-Julien]{arpit2017icml}
Arpit, D., Jastrzebski, S., Ballas, N., Krueger, D., Bengio, E., Kanwal, M.~S.,
  Maharaj, T., Fischer, A., Courville, A., Bengio, Y., and Lacoste-Julien, S.
\newblock A closer look at memorization in deep networks.
\newblock In \emph{ICML}, 2017.

\bibitem[Belkin et~al.(2018)Belkin, Hsu, and Mitra]{belkin2018neurips}
Belkin, M., Hsu, D.~J., and Mitra, P.
\newblock Overfitting or perfect fitting? {R}isk bounds for classification and
  regression rules that interpolate.
\newblock In \emph{NeurIPS}, 2018.

\bibitem[Belkin et~al.(2019)Belkin, Hsu, Ma, and Mandal]{belkin2019pnas}
Belkin, M., Hsu, D., Ma, S., and Mandal, S.
\newblock Reconciling modern machine-learning practice and the classical
  bias–variance trade-off.
\newblock \emph{PNAS}, 116:\penalty0 15850--15854, 2019.

\bibitem[Berthelot et~al.(2019)Berthelot, Carlini, Goodfellow, Papernot,
  Oliver, and Raffel]{Berthelot2019neurips}
Berthelot, D., Carlini, N., Goodfellow, I., Papernot, N., Oliver, A., and
  Raffel, C.~A.
\newblock Mix{M}atch: A holistic approach to semi-supervised learning.
\newblock In \emph{NeurIPS}, 2019.

\bibitem[Bishop(1995)]{bishop1995icann}
Bishop, C.~M.
\newblock Regularization and complexity control in feed-forward networks.
\newblock In \emph{ICANN}, 1995.

\bibitem[Bishop(2011)]{bishop2011book}
Bishop, C.~M.
\newblock \emph{Pattern Recognition and Machine Learning (Information Science
  and Statistics)}.
\newblock Springer, 2011.

\bibitem[Caruana et~al.(2000)Caruana, Lawrence, and Giles]{caruana2000neurips}
Caruana, R., Lawrence, S., and Giles, C.~L.
\newblock Overfitting in neural nets: Backpropagation, conjugate gradient, and
  early stopping.
\newblock In \emph{NeurIPS}, 2000.

\bibitem[Chaudhari et~al.(2017)Chaudhari, Choromanska, Soatto, LeCun, Baldassi,
  Borgs, Chayes, Sagun, and Zecchina]{chaudhari2017iclr}
Chaudhari, P., Choromanska, A., Soatto, S., LeCun, Y., Baldassi, C., Borgs, C.,
  Chayes, J., Sagun, L., and Zecchina, R.
\newblock Entropy-{SGD}: Biasing gradient descent into wide valleys.
\newblock In \emph{ICLR}, 2017.

\bibitem[Cid-Sueiro et~al.(2014)Cid-Sueiro, Garc\'{i}a-Garc\'{i}a, and
  Santos-Rodr\'{i}guez]{jesus14ecml}
Cid-Sueiro, J., Garc\'{i}a-Garc\'{i}a, D., and Santos-Rodr\'{i}guez, R.
\newblock Consistency of losses for learning from weak labels.
\newblock In \emph{ECML-PKDD}, 2014.

\bibitem[Clanuwat et~al.(2018)Clanuwat, Bober-Irizar, Kitamoto, Lamb, Yamamoto,
  and Ha]{clanuwat18neurips}
Clanuwat, T., Bober-Irizar, M., Kitamoto, A., Lamb, A., Yamamoto, K., and Ha,
  D.
\newblock Deep learning for classical {J}apanese literature.
\newblock In \emph{NeurIPS Workshop on Machine Learning for Creativity and
  Design}, 2018.

\bibitem[{du Plessis} et~al.(2014){du Plessis}, Niu, and
  Sugiyama]{christo14neurips}
{du Plessis}, M.~C., Niu, G., and Sugiyama, M.
\newblock Analysis of learning from positive and unlabeled data.
\newblock In \emph{NeurIPS}, 2014.

\bibitem[{du Plessis} et~al.(2015){du Plessis}, Niu, and
  Sugiyama]{christo15icml}
{du Plessis}, M.~C., Niu, G., and Sugiyama, M.
\newblock Convex formulation for learning from positive and unlabeled data.
\newblock In \emph{ICML}, 2015.

\bibitem[Goodfellow et~al.(2016)Goodfellow, Bengio, and
  Courville]{goodfellow2016book}
Goodfellow, I., Bengio, Y., and Courville, A.
\newblock \emph{Deep Learning}.
\newblock MIT Press, 2016.

\bibitem[Guo et~al.(2017)Guo, Pleiss, Sun, and Weinberger]{guo2017icml}
Guo, C., Pleiss, G., Sun, Y., and Weinberger, K.~Q.
\newblock On calibration of modern neural networks.
\newblock In \emph{ICML}, 2017.

\bibitem[Guo et~al.(2019)Guo, Mao, and Zhang]{guo2019arxiv}
Guo, H., Mao, Y., and Zhang, R.
\newblock Augmenting data with mixup for sentence classification: An empirical
  study.
\newblock In \emph{arXiv:1905.08941}, 2019.

\bibitem[Han et~al.(2020)Han, Niu, Yu, Yao, Xu, Tsang, and
  Sugiyama]{han2020icml}
Han, B., Niu, G., Yu, X., Yao, Q., Xu, M., Tsang, I.~W., and Sugiyama, M.
\newblock Sigua: Forgetting may make learning with noisy labels more robust.
\newblock In \emph{ICML}, 2020.

\bibitem[Hanson \& Pratt(1988)Hanson and Pratt]{hanson1988nips}
Hanson, S.~J. and Pratt, L.~Y.
\newblock Comparing biases for minimal network construction with
  back-propagation.
\newblock In \emph{NeurIPS}, 1988.

\bibitem[He et~al.(2016)He, Zhang, Ren, and Sun]{he2016cvpr}
He, K., Zhang, X., Ren, S., and Sun, J.
\newblock Deep residual learning for image recognition.
\newblock In \emph{CVPR}, 2016.

\bibitem[Hochreiter \& Schmidhuber(1997)Hochreiter and
  Schmidhuber]{schmidhuber1997neuco}
Hochreiter, S. and Schmidhuber, J.
\newblock Flat minima.
\newblock \emph{Neural Computation}, 9:\penalty0 1--42, 1997.

\bibitem[Idelbayev(2020)]{akamasterrepo}
Idelbayev, Y.
\newblock Proper {ResNet} implementation for {CIFAR10/CIFAR100} in {PyTorch}.
\newblock \url{https//github.com/akamaster/pytorch_resnet_cifar10}, 2020.
\newblock Accessed: 2020-05-31.

\bibitem[Ioffe \& Szegedy(2015)Ioffe and Szegedy]{ioffe2015icml}
Ioffe, S. and Szegedy, C.
\newblock Batch normalization: Accelerating deep network training by reducing
  internal covariate shift.
\newblock In \emph{ICML}, 2015.

\bibitem[Ishida et~al.(2019)Ishida, Niu, Menon, and Sugiyama]{ishida2019icml}
Ishida, T., Niu, G., Menon, A.~K., and Sugiyama, M.
\newblock Complementary-label learning for arbitrary losses and models.
\newblock In \emph{ICML}, 2019.

\bibitem[Keskar et~al.(2017)Keskar, Mudigere, Nocedal, Smelyanskiy, and
  Tang]{keskar2017iclr}
Keskar, N.~S., Mudigere, D., Nocedal, J., Smelyanskiy, M., and Tang, P. T.~P.
\newblock On large-batch training for deep learning: Generalization gap and
  sharp minima.
\newblock In \emph{ICLR}, 2017.

\bibitem[Kingma \& Ba(2015)Kingma and Ba]{kingma15iclr}
Kingma, D.~P. and Ba, J.~L.
\newblock Adam: A method for stochastic optimization.
\newblock In \emph{ICLR}, 2015.

\bibitem[Kiryo et~al.(2017)Kiryo, Niu, {du Plessis}, and Sugiyama]{kiryo17nips}
Kiryo, R., Niu, G., {du Plessis}, M.~C., and Sugiyama, M.
\newblock Positive-unlabeled learning with non-negative risk estimator.
\newblock In \emph{NeurIPS}, 2017.

\bibitem[Kolesnikov et~al.(2020)Kolesnikov, Beyer, Zhai, Puigcerver, Yung,
  Gelly, and Houlsby]{kolesnikov2020arxiv}
Kolesnikov, A., Beyer, L., Zhai, X., Puigcerver, J., Yung, J., Gelly, S., and
  Houlsby, N.
\newblock Big transfer ({B}i{T}): {G}eneral visual representation learning.
\newblock In \emph{arXiv:1912.11370v3}, 2020.

\bibitem[Krogh \& Hertz(1992)Krogh and Hertz]{krogh1992neurips}
Krogh, A. and Hertz, J.~A.
\newblock A simple weight decay can improve generalization.
\newblock In \emph{NeurIPS}, 1992.

\bibitem[Lecun et~al.(1998)Lecun, Bottou, Bengio, and
  Haffner]{Lecun98gradient-basedlearning}
Lecun, Y., Bottou, L., Bengio, Y., and Haffner, P.
\newblock Gradient-based learning applied to document recognition.
\newblock In \emph{Proceedings of the IEEE}, pp.\  2278--2324, 1998.

\bibitem[Li et~al.(2018)Li, Xu, Taylor, Studer, and Goldstein]{li2018neurips}
Li, H., Xu, Z., Taylor, G., Studer, C., and Goldstein, T.
\newblock Visualizing the loss landscape of neural nets.
\newblock In \emph{NeurIPS}, 2018.

\bibitem[Loshchilov \& Hutter(2019)Loshchilov and Hutter]{loschilov2019iclr}
Loshchilov, I. and Hutter, F.
\newblock Decoupled weight decay regularization.
\newblock In \emph{ICLR}, 2019.

\bibitem[Lu et~al.(2020)Lu, Zhang, Niu, and Sugiyama]{lu2020aistats}
Lu, N., Zhang, T., Niu, G., and Sugiyama, M.
\newblock Mitigating overfitting in supervised classification from two
  unlabeled datasets: A consistent risk correction approach.
\newblock In \emph{AISTATS}, 2020.

\bibitem[Morgan \& Bourlard(1990)Morgan and Bourlard]{morgan1990nips}
Morgan, N. and Bourlard, H.
\newblock Generalization and parameter estimation in feedforward nets: Some
  experiments.
\newblock In \emph{NeurIPS}, 1990.

\bibitem[Nair \& Hinton(2010)Nair and Hinton]{nair10icml}
Nair, V. and Hinton, G.
\newblock Rectified linear units improve restricted boltzmann machines.
\newblock In \emph{ICML}, 2010.

\bibitem[Nakkiran et~al.(2019)Nakkiran, Kaplun, Kalimeris, Yang, Edelman,
  Zhang, and Barak]{nakkiran2019neurips}
Nakkiran, P., Kaplun, G., Kalimeris, D., Yang, T., Edelman, B.~L., Zhang, F.,
  and Barak, B.
\newblock {SGD} on neural networks learns functions of increasing complexity.
\newblock In \emph{NeurIPS}, 2019.

\bibitem[Nakkiran et~al.(2020)Nakkiran, Kaplun, Bansal, Yang, Barak, and
  Sutskever]{nakkiran2020iclr}
Nakkiran, P., Kaplun, G., Bansal, Y., Yang, T., Barak, B., and Sutskever, I.
\newblock Deep double descent: Where bigger models and more data hurt.
\newblock In \emph{ICLR}, 2020.

\bibitem[Natarajan et~al.(2013)Natarajan, Dhillon, Ravikumar, and
  Tewari]{natarajan2013nips}
Natarajan, N., Dhillon, I.~S., Ravikumar, P.~K., and Tewari, A.
\newblock Learning with noisy labels.
\newblock In \emph{NeurIPS}, 2013.

\bibitem[Netzer et~al.(2011)Netzer, Wang, Coates, Bissacco, Wu, and
  Ng]{netzer2011neurips}
Netzer, Y., Wang, T., Coates, A., Bissacco, A., Wu, B., and Ng, A.~Y.
\newblock Reading digits in natural images with unsupervised feature learning.
\newblock In \emph{NeurIPS Workshop on Deep Learning and Unsupervised Feature
  Learning}, 2011.

\bibitem[Ng(1997)]{ng1997icml}
Ng, A.~Y.
\newblock Preventing ``overfitting'' of cross-validation data.
\newblock In \emph{ICML}, 1997.

\bibitem[Paszke et~al.(2019)Paszke, Gross, Massa, Lerer, Bradbury, Chanan,
  Killeen, Lin, Gimelshein, Antiga, Desmaison, Kopf, Yang, DeVito, Raison,
  Tejani, Chilamkurthy, Steiner, Fang, Bai, and Chintala]{paszke2019neurips}
Paszke, A., Gross, S., Massa, F., Lerer, A., Bradbury, J., Chanan, G., Killeen,
  T., Lin, Z., Gimelshein, N., Antiga, L., Desmaison, A., Kopf, A., Yang, E.,
  DeVito, Z., Raison, M., Tejani, A., Chilamkurthy, S., Steiner, B., Fang, L.,
  Bai, J., and Chintala, S.
\newblock Py{T}orch: {A}n imperative style, high-performance deep learning
  library.
\newblock In \emph{NeurIPS}, 2019.

\bibitem[Patrini et~al.(2017)Patrini, Rozza, Menon, Nock, and
  Qu]{patrini17cvpr}
Patrini, G., Rozza, A., Menon, A.~K., Nock, R., and Qu, L.
\newblock Making deep neural networks robust to label noise: A loss correction
  approach.
\newblock In \emph{CVPR}, 2017.

\bibitem[Robbins \& Monro(1951)Robbins and Monro]{robbins1951}
Robbins, H. and Monro, S.
\newblock A stochastic approximation method.
\newblock \emph{Annals of Mathematical Statistics}, 22:\penalty0 400--407,
  1951.

\bibitem[Roelofs et~al.(2019)Roelofs, Shankar, Recht, Fridovich-Keil, Hardt,
  Miller, and Schmidt]{roelofs2019neurips}
Roelofs, R., Shankar, V., Recht, B., Fridovich-Keil, S., Hardt, M., Miller, J.,
  and Schmidt, L.
\newblock A meta-analysis of overfitting in machine learning.
\newblock In \emph{NeurIPS}, 2019.

\bibitem[Shorten \& Khoshgoftaar(2019)Shorten and Khoshgoftaar]{shorten2019jbg}
Shorten, C. and Khoshgoftaar, T.~M.
\newblock A survey on image data augmentation for deep learning.
\newblock \emph{Journal of Big Data}, 6, 2019.

\bibitem[Srivastava et~al.(2014)Srivastava, Hinton, Krizhevsky, Sutskever, and
  Salakhutdinov]{srivastava2014jmlr}
Srivastava, N., Hinton, G., Krizhevsky, A., Sutskever, I., and Salakhutdinov,
  R.
\newblock Dropout: A simple way to prevent neural networks from overfitting.
\newblock \emph{Journal of Machine Learning Research}, 15:\penalty0 1929--1958,
  2014.

\bibitem[Sugiyama(2015)]{sugiyama2015introduction}
Sugiyama, M.
\newblock \emph{Introduction to statistical machine learning}.
\newblock Morgan Kaufmann, 2015.

\bibitem[Szegedy et~al.(2016)Szegedy, Vanhoucke, Ioffe, and
  Shlens]{szegedy2016cvpr}
Szegedy, C., Vanhoucke, V., Ioffe, S., and Shlens, J.
\newblock Rethinking the inception architecture for computer vision.
\newblock In \emph{CVPR}, 2016.

\bibitem[Thulasidasan et~al.(2019)Thulasidasan, Chennupati, Bilmes,
  Bhattacharya, and Michalak]{thulasidasan2019neurips}
Thulasidasan, S., Chennupati, G., Bilmes, J., Bhattacharya, T., and Michalak,
  S.
\newblock On mixup training: Improved calibration and predictive uncertainty
  for deep neural networks.
\newblock In \emph{NeurIPS}, 2019.

\bibitem[Tibshirani(1996)]{tibshirani1996jrss}
Tibshirani, R.
\newblock Regression shrinkage and selection via the lasso.
\newblock \emph{Journal of the Royal Statistical Society: Series B
  (Methodological)}, 58:\penalty0 267--288, 1996.

\bibitem[Tikhonov(1943)]{tikhonov1943sssr}
Tikhonov, A.~N.
\newblock On the stability of inverse problems.
\newblock \emph{Doklady Akademii Nauk SSSR}, 39:\penalty0 195--198, 1943.

\bibitem[Tikhonov \& Arsenin(1977)Tikhonov and Arsenin]{tikhonov1977book}
Tikhonov, A.~N. and Arsenin, V.~Y.
\newblock \emph{Solutions of Ill Posed Problems}.
\newblock Winston, 1977.

\bibitem[Torralba et~al.(2008)Torralba, Fergus, and Freeman]{Torralba08pami}
Torralba, A., Fergus, R., and Freeman, W.~T.
\newblock 80 million tiny images: {A} large data set for nonparametric object
  and scene recognition.
\newblock In \emph{IEEE Trans. PAMI}, 2008.

\bibitem[van Rooyen \& Williamson(2018)van Rooyen and
  Williamson]{rooyen2018jmlr}
van Rooyen, B. and Williamson, R.~C.
\newblock A theory of learning with corrupted labels.
\newblock \emph{Journal of Machine Learning Research}, 18:\penalty0 1--50,
  2018.

\bibitem[Verma et~al.(2019)Verma, Lamb, Kannala, Bengio, and
  Lopez-Paz]{verma2019ijcal}
Verma, V., Lamb, A., Kannala, J., Bengio, Y., and Lopez-Paz, D.
\newblock Interpolation consistency training for semi-supervised learning.
\newblock In \emph{IJCAI}, 2019.

\bibitem[Wager et~al.(2013)Wager, Wang, and Liang]{wager2013nips}
Wager, S., Wang, S., and Liang, P.
\newblock Dropout training as adaptive regularization.
\newblock In \emph{NeurIPS}, 2013.

\bibitem[Werpachowski et~al.(2019)Werpachowski, György, and
  Szepesvári]{wepachowski2019neurips}
Werpachowski, R., György, A., and Szepesvári, C.
\newblock Detecting overfitting via adversarial examples.
\newblock In \emph{NeurIPS}, 2019.

\bibitem[Zhang et~al.(2017)Zhang, Bengio, Hardt, Recht, and
  Vinyals]{zhang2017iclr}
Zhang, C., Bengio, S., Hardt, M., Recht, B., and Vinyals, O.
\newblock Understanding deep learning requires rethinking generalization.
\newblock In \emph{ICLR}, 2017.

\bibitem[Zhang et~al.(2018)Zhang, Cisse, Dauphin, and Lopez-Paz]{zhang2018iclr}
Zhang, H., Cisse, M., Dauphin, Y.~N., and Lopez-Paz, D.
\newblock {m}ixup: Beyond empirical risk minimization.
\newblock In \emph{ICLR}, 2018.

\bibitem[Zhang(2004)]{zhang2004as}
Zhang, T.
\newblock Statistical behavior and consistency of classification methods based
  on convex risk minimization.
\newblock \emph{The Annals of Statistics}, 32:\penalty0 56--85, 2004.

\end{thebibliography}
\bibliographystyle{icml2020}


\newpage
\onecolumn
\appendix


\section{Proof of Theorem~\ref{theorem:mse}}
\label{appsec:mse_proof}

\begin{proof}
If the flood level is $b$, then the proposed flooding estimator is 
\begin{equation}
\widetilde{R}(\bmg) = |\widehat{R}(\bmg)-b|+b.
\end{equation}
Since the absolute operator can be expressed with a max operator with $\max(a,b)=\frac{a+b+|a-b|}{2}$, the proposed estimator can be re-expressed as
\begin{align}
    \widetilde{R}(\bmg) = 2\max(\widehat{R}(\bmg),b)-\widehat{R}(\bmg) = A - \widehat{R}(\bmg).
\end{align}
For convenience, we used $A = 2\max(\widehat{R}(\bmg),b)$.
From the definition of MSE,
\begin{equation}
\mathrm{MSE}(\widehat{R}(\bmg)) = \mathbb{E}[(\widehat{R}(\bmg)-R(\bmg))^2],
\end{equation}
and
\begin{align}
\mathrm{MSE}(\widetilde{R}(\bmg))&= \mathbb{E}[(\widetilde{R}(\bmg)-R(\bmg))^2]\\
&=\mathbb{E}[(A-\widehat{R}(\bmg)-R(\bmg))^2)]\\
&=\mathbb{E}[A^2] -2\mathbb{E}[A(\widehat{R}(\bmg)+R(\bmg))]+\mathbb{E}[(\widehat{R}(\bmg)+R(\bmg))^2].
\end{align}
We are interested in the sign of
\begin{align}
\mathrm{MSE}(\widehat{R}(\bmg)) - \mathrm{MSE}(\widetilde{R}(\bmg)) &=
\mathbb{E}[-4\widehat{R}(\bmg)R(\bmg)-A^2+2A(\widehat{R}(\bmg)+R(\bmg))].
\end{align}
Define the inside of the expectation as $B=-4\widehat{R}(\bmg)R(\bmg)-A^2+2A(\widehat{R}(\bmg)+R(\bmg))$.
$B$ can be divided into two cases, depending on the outcome of the max operator:
\begin{align}
B =&
  \begin{cases}
    -4\widehat{R}(\bmg)R(\bmg) -4\widehat{R}(\bmg)^2+4\widehat{R}(\bmg)(\widehat{R}(\bmg)+R(\bmg)) \quad& \text{if}\quad \widehat{R}(\bmg)\geq b \\
    -4\widehat{R}(\bmg)R(\bmg)-4b^2+4b(\widehat{R}(\bmg)+R(\bmg)) & \text{if}\quad \widehat{R}(\bmg)<b
  \end{cases}\\
  =&
    \begin{cases}
    0 \quad& \text{if}\quad \widehat{R}(\bmg)\geq b \\
    -4(b-\widehat{R}(\bmg))(b-R(\bmg)) & \text{if}\quad \widehat{R}(\bmg)<b.
  \end{cases}
\end{align}
In the latter case, $B$ becomes positive when $\widehat{R}(\bmg)<b<R(\bmg)$.

Therefore, if $b \le R(\bmg)$, we have
\begin{align}
\operatorname{MSE}(\widehat{R}(\bmg)) - \operatorname{MSE}(\widetilde{R}(\bmg))
&= \mathbb{E}[B \mid \widehat{R}(\bmg) \ge b] \Pr[\widehat{R}(\bmg) \ge b]
+ \mathbb{E}[B \mid \widehat{R}(\bmg) < b] \Pr[\widehat{R}(\bmg) < b]\\
&= \mathbb{E}[B \mid \widehat{R}(\bmg) < b] \Pr[\widehat{R}(\bmg) < b]\\
&\ge 0.
\end{align}
Furthermore, if $b < R(\bmg)$ and $\Pr[\widehat{R}(\bmg) < b] > 0$, we have
\begin{align}
\operatorname{MSE}(\widehat{R}(\bmg)) - \operatorname{MSE}(\widetilde{R}(\bmg))
= \mathbb{E}[B \mid \widehat{R}(\bmg) < b] \Pr[\widehat{R}(\bmg) < b]
> 0.
\end{align}
\end{proof}

\section{Benchmark Datasets}
\label{appsec:bench_datasets}
In the experiments in Section~\ref{sec:benchmark}, we use 6 image benchmark datasets explained below.
\begin{itemize}
\item MNIST\footnote{\url{http://yann.lecun.com/exdb/mnist/}} \citep{Lecun98gradient-basedlearning} is a 10 class dataset of handwritten digits: $1,2\ldots, 9$ and $0$.  Each sample is a $28\times28$ grayscale image.  The number of training and test samples are 60,000 and 10,000, respectively.
\item Kuzushiji-MNIST\footnote{\url{https://github.com/rois-codh/kmnist}} \citep{clanuwat18neurips} is a 10 class dataset of cursive Japanese (``Kuzushiji'') characters.  Each sample is a $28\times28$ grayscale image.  The number of training and test samples are 60,000 and 10,000, respectively.
\item SVHN\footnote{\url{http://ufldl.stanford.edu/housenumbers/}} \citep{netzer2011neurips} is a 10 class dataset of house numbers from Google Street View images, in $32 \times 32 \times 3$ RGB format.
73257 digits are for training and 26032 digits are for testing.
\item CIFAR-10\footnote{\url{https://www.cs.toronto.edu/~kriz/cifar.html}} is a 10 class dataset of various objects: airplane, automobile, bird, cat, deer, dog, frog, horse, ship, and truck.
Each sample is a colored image in $32\times32\times3$ RGB format.
It is a subset of the 80 million tiny images dataset \citep{Torralba08pami}.
There are 6,000 images per class, where 5,000 are for training and 1,000 are for test.
\item CIFAR-100\footnote{\url{https://www.cs.toronto.edu/~kriz/cifar.html}} is a 100 class dataset of various objects.
Each class has 600 samples, where 500 samples are for training and 100 samples are for test.
This is also a subset of the 80 million tiny images dataset \citep{Torralba08pami}.
\end{itemize}

\section{Learning Curves}
\label{appsec:learning_curves}

\begin{figure}[ht]
\centering
\subcaptionbox{MNIST, MLP}{\includegraphics[width=\columnwidth*13/40]{images/lr_curves/MNIST_mlp_model_2hl_DA_False_LRdecay_False_0730_mnist_loss.pdf}}
\subcaptionbox{MNIST, MLP\&BN}{\includegraphics[width=\columnwidth*13/40]{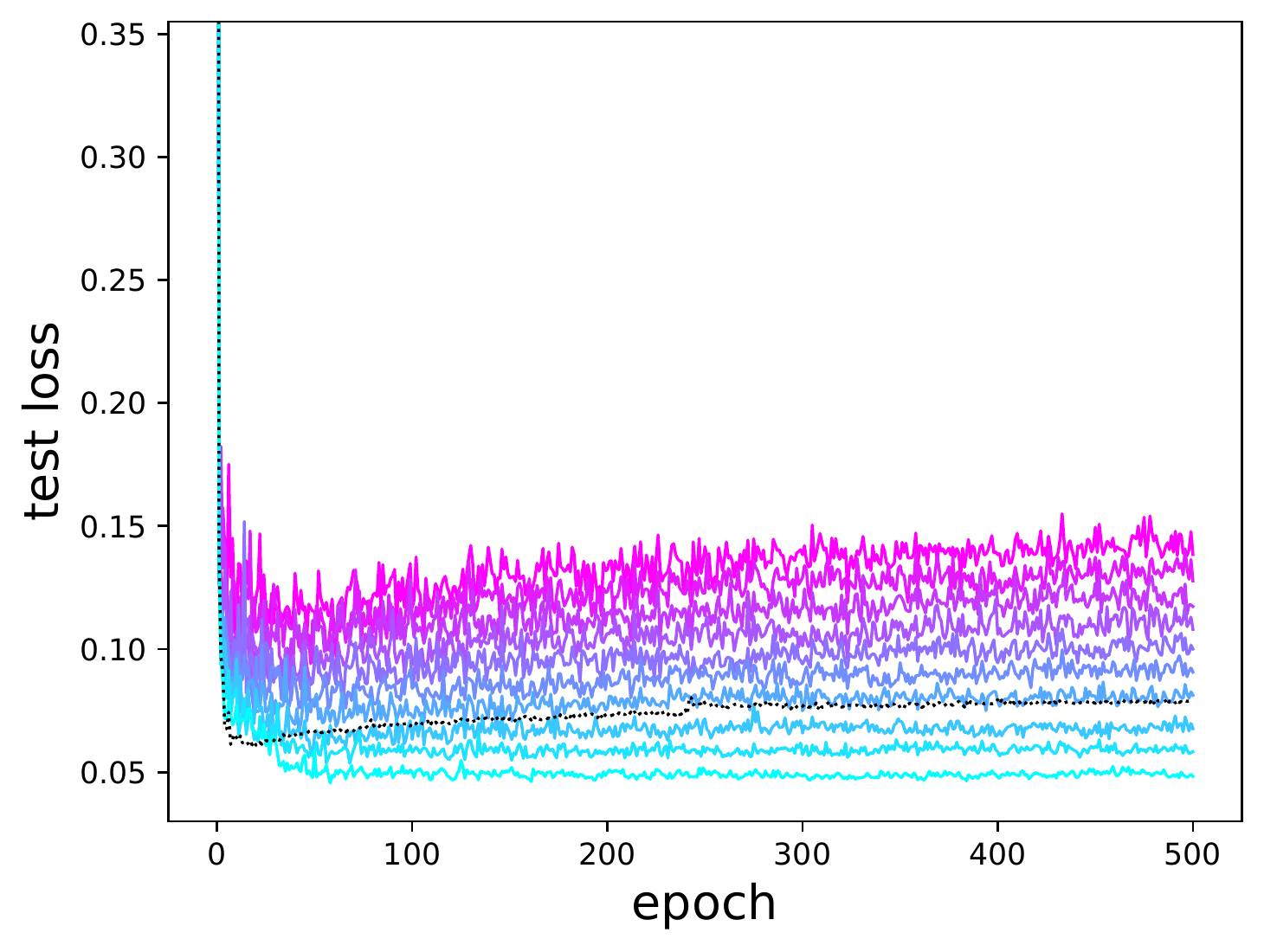}}
\subcaptionbox{KMNIST, MLP}{\includegraphics[width=\columnwidth*13/40]{images/lr_curves/KMNIST_mlp_model_2hl_DA_False_LRdecay_False_0730_kmnist_loss.pdf}}
\subcaptionbox{KMNIST, MLP\&BN}{\includegraphics[width=\columnwidth*13/40]{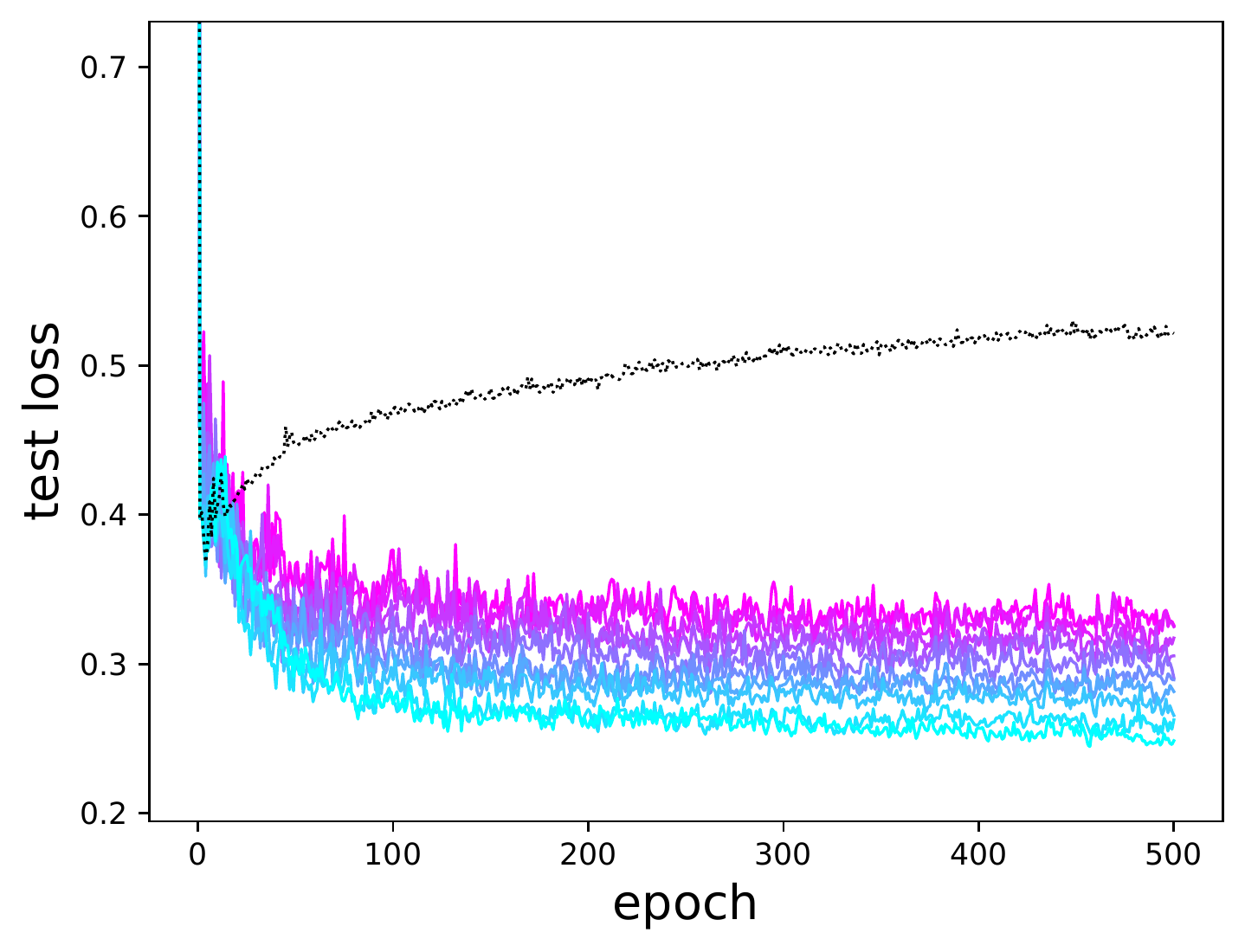}}
\subcaptionbox{SVHN, ResNet18}{\includegraphics[width=\columnwidth*13/40]{images/lr_curves/SVHN_resnet18_DA_False_LRdecay_False_0802_svhn_resnet18_loss.pdf}}
\subcaptionbox{C10, ResNet44}{\includegraphics[width=\columnwidth*13/40]{images/lr_curves/CIFAR10_resnet44_DA_False_LRdecay_False_0802_c10_resnet44_loss.pdf}}
\subcaptionbox{C10, ResNet44, DA \& LRD}{\includegraphics[width=\columnwidth*13/40]{images/lr_curves/CIFAR10_resnet44_DA_True_LRdecay_True_0809_c10_resnet44_da_lrd_loss.pdf}}
\subcaptionbox{C100, ResNet44}{\includegraphics[width=\columnwidth*13/40]{images/lr_curves/CIFAR100_resnet44_DA_False_LRdecay_False_0802_c100_resnet44_loss.pdf}}
\subcaptionbox{C100, ResNet44, DA \& LRD}{\includegraphics[width=\columnwidth*13/40]{images/lr_curves/CIFAR100_resnet44_DA_True_LRdecay_True_0809_c100_resnet44_da_lrd_loss.pdf}}
\includegraphics[width=\columnwidth*1/2]{images/lr_curves/colorbar.pdf}
\caption{
Learning curves of the test loss showing that adding flooding leads to lower test loss.
The black dotted line shows the baseline without flooding. The colored lines show the learning curves with flooding for different flooding levels.  We show the learning curves for $b\in \{0.01, 0.02, \ldots, 0.10\}$.
KMNIST is Kuzushiji-MNIST, C10 is CIFAR-10, and C100 is CIFAR-100.
MLP, BN, DA, and LRD stand for multi-layer perceptron, batch normalization, data augmentation and learning rate decay, respectively.
}
\label{appfig:learning_curves_loss}
\end{figure}

In Fig.~\ref{appfig:learning_curves_loss}, we show more results of learning curves of the test loss.
Note that Figure~\ref{fig:overfittings-c} shows the learning curves for the first 80 epochs for CIFAR-10 without flooding with ResNet-18.
Figure~\ref{fig:overfittings-d} shows the learning curves with flooding, when the flood level is $0.18$ also with ResNet-18.
Other details follow the experiments in Section~\ref{sec:benchmark}.

\section{Additional Figures for Section~\ref{subsec:gradients} and Section~\ref{subsec:flatness}}
\label{appsec:more_figs}
\begin{figure*}[ht]
\centering
\subcaptionbox{MNIST, x:train}{\includegraphics[width=\columnwidth*19/80]{images/gradients/MNIST_Loss_train.pdf}}
\subcaptionbox{MNIST, x:test}{\includegraphics[width=\columnwidth*19/80]{images/gradients/MNIST_Loss_test.pdf}}
\subcaptionbox{KMNIST, x:train}{\includegraphics[width=\columnwidth*19/80]{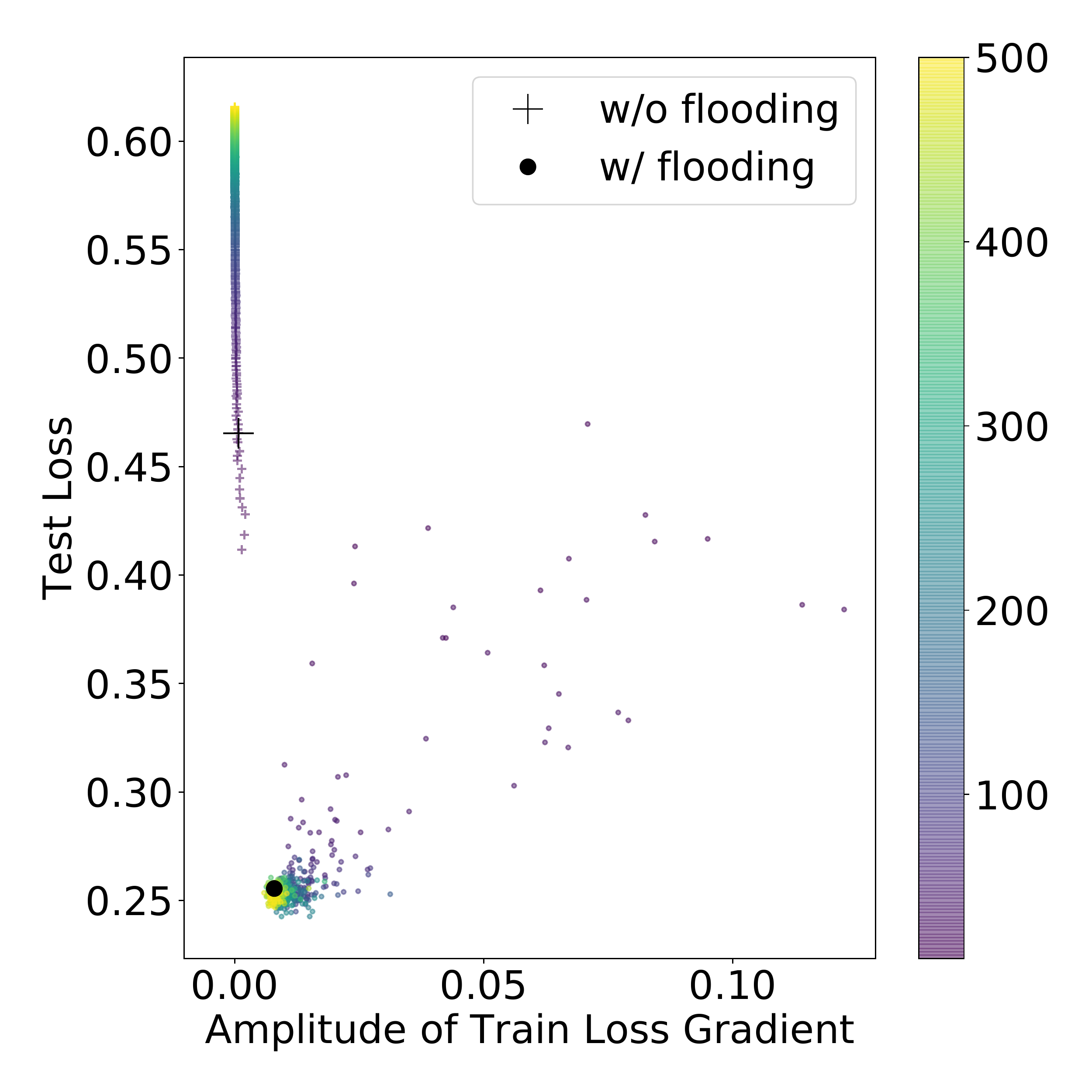}}
\subcaptionbox{KMNIST, x:test}{\includegraphics[width=\columnwidth*19/80]{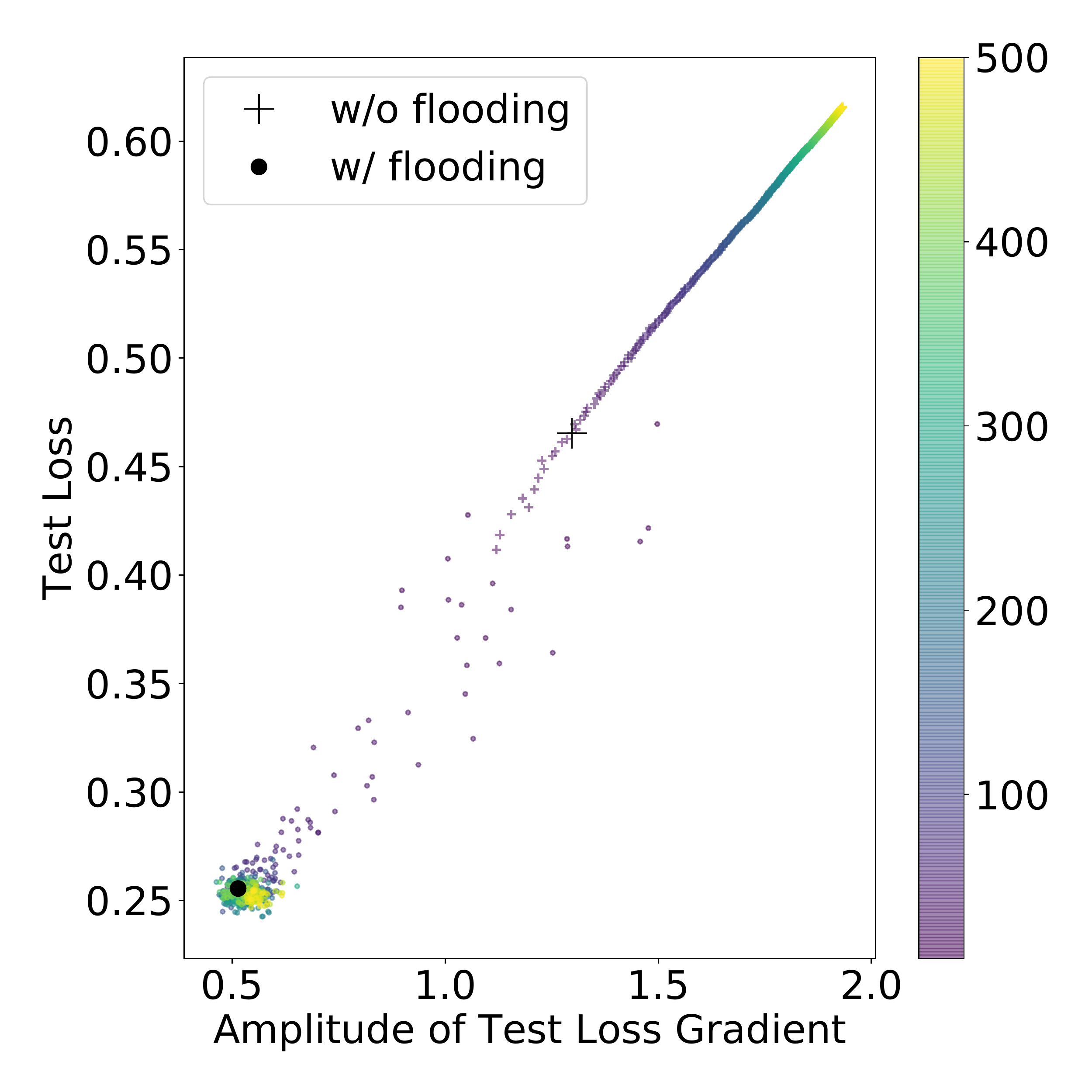}}
\subcaptionbox{SVHN, x:train}{\includegraphics[width=\columnwidth*19/80]{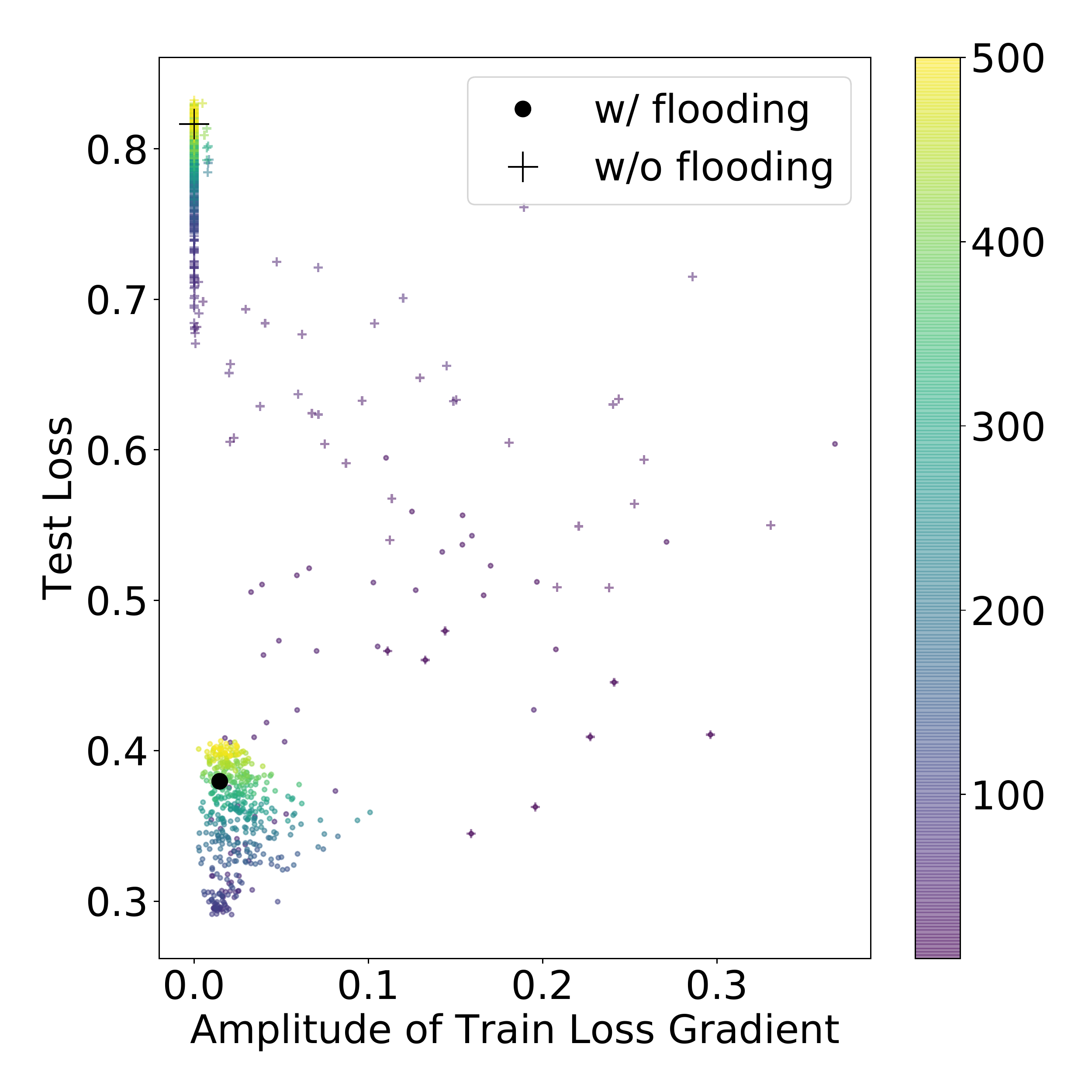}}
\subcaptionbox{SVHN, x:test}{\includegraphics[width=\columnwidth*19/80]{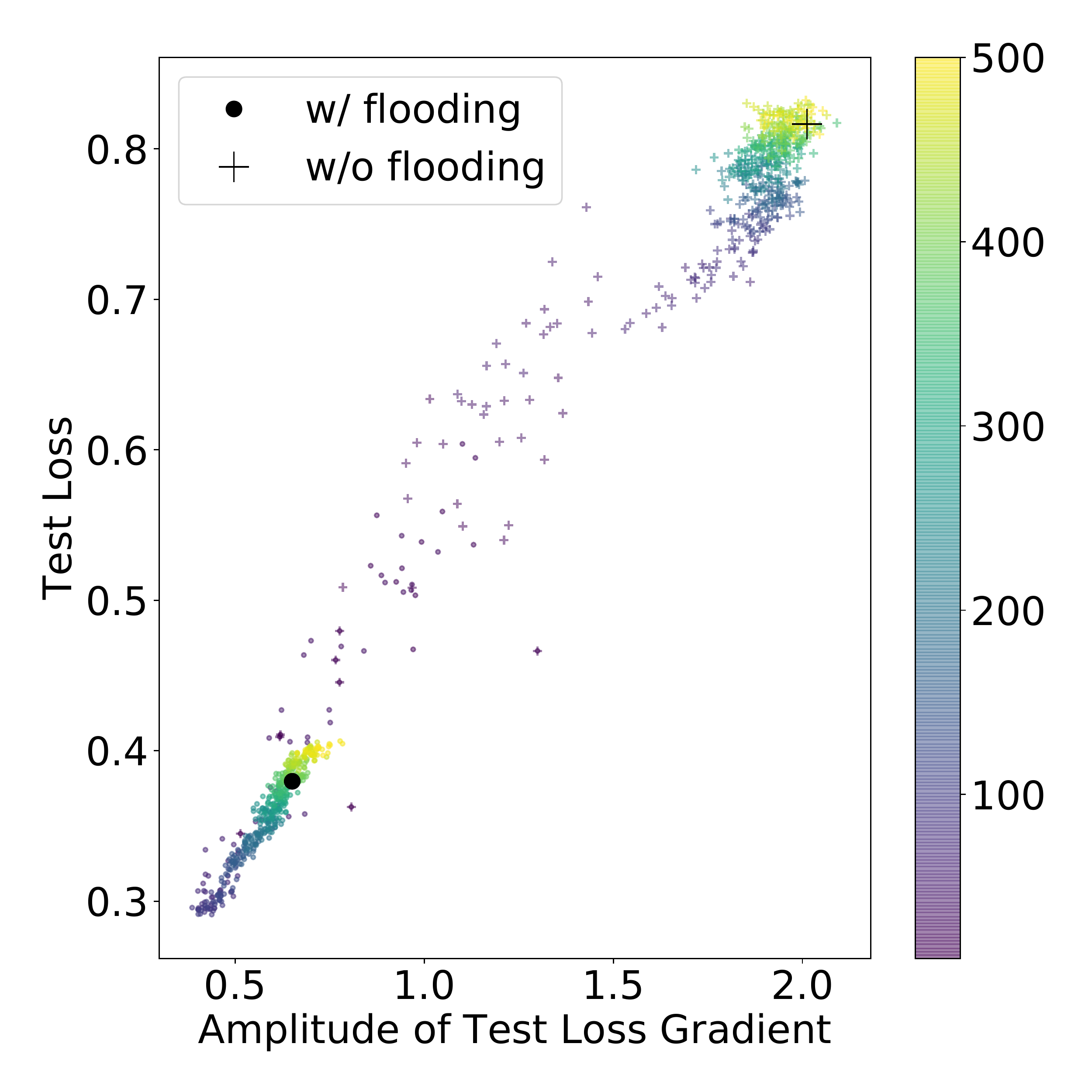}}
\subcaptionbox{CIFAR-10, x:train}{\includegraphics[width=\columnwidth*19/80]{images/gradients/CIFAR10_Loss_train.pdf}}
\subcaptionbox{CIFAR-10, x:test}{\includegraphics[width=\columnwidth*19/80]{images/gradients/CIFAR10_Loss_test.pdf}}
\subcaptionbox{CIFAR-100, x:train}{\includegraphics[width=\columnwidth*19/80]{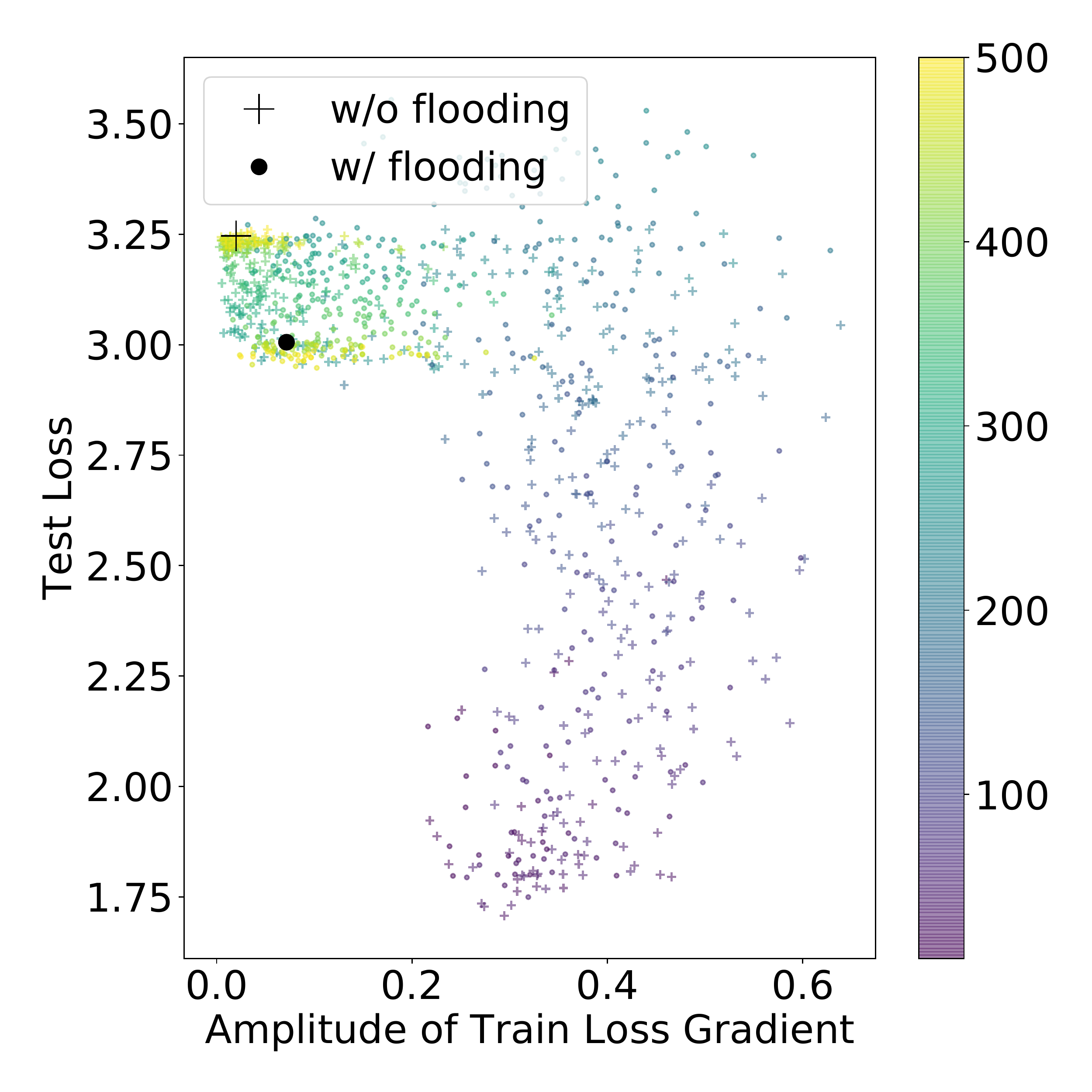}}
\subcaptionbox{CIFAR-100, x:test}{\includegraphics[width=\columnwidth*19/80]{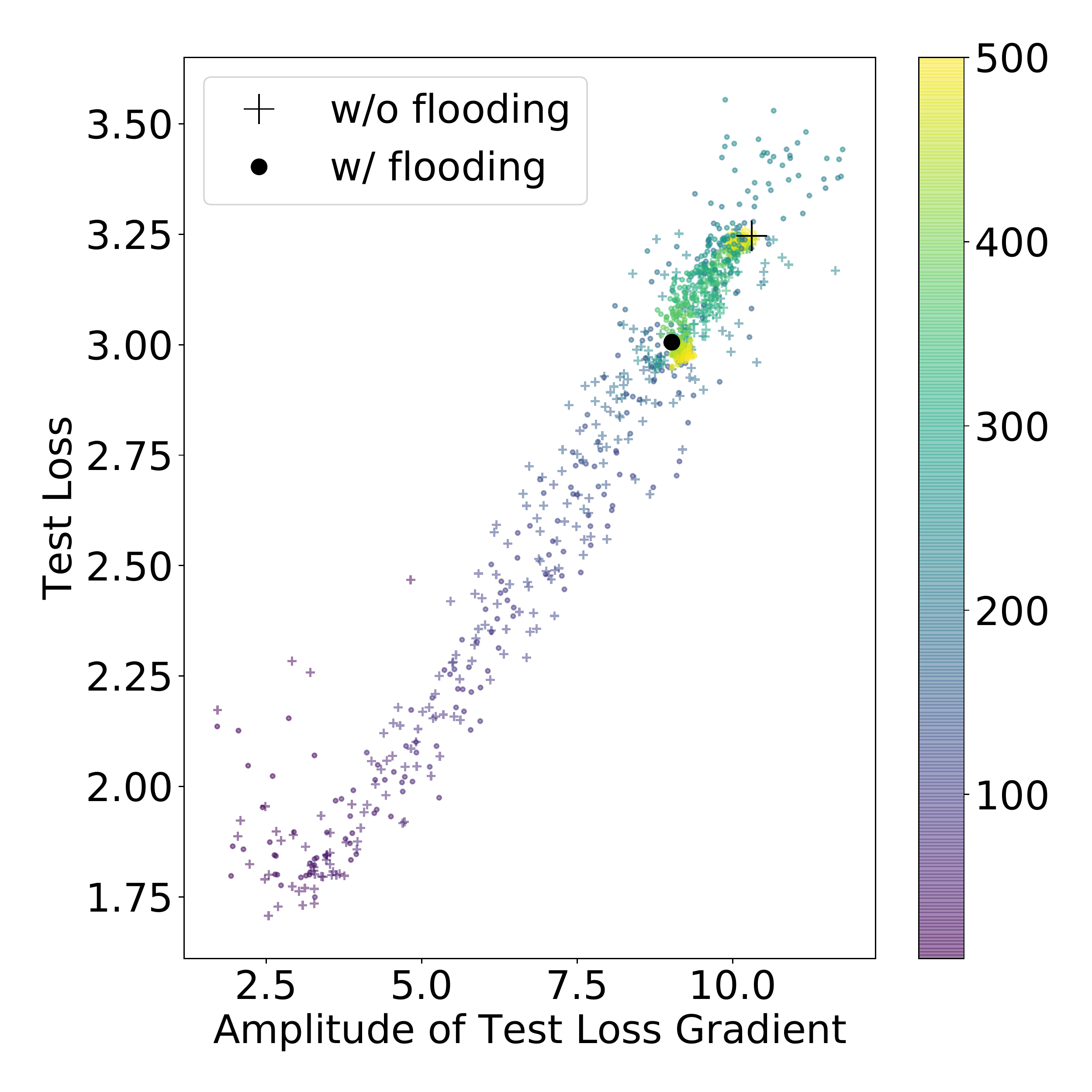}}
\caption{
Relationship between test loss and amplitude of gradient (with training or test loss).
Each point (``$\circ$'' or ``$+$'') in the figures corresponds to a single model at a certain epoch.
We remove the first 10 epochs and plot the rest.
``$\circ$'' is used for the method with flooding  and ``$+$'' is used for the method without flooding.
The large black ``$\circ$'' and ``$+$'' show the epochs with early stopping.
The color becomes lighter (purple $\rightarrow$ yellow) as the training proceeds.
}
\label{appfig:performance_gradients}
\end{figure*}

\begin{figure*}[htb]
\centering
\subcaptionbox{MNIST (train)}{\includegraphics[width=\columnwidth*19/80]{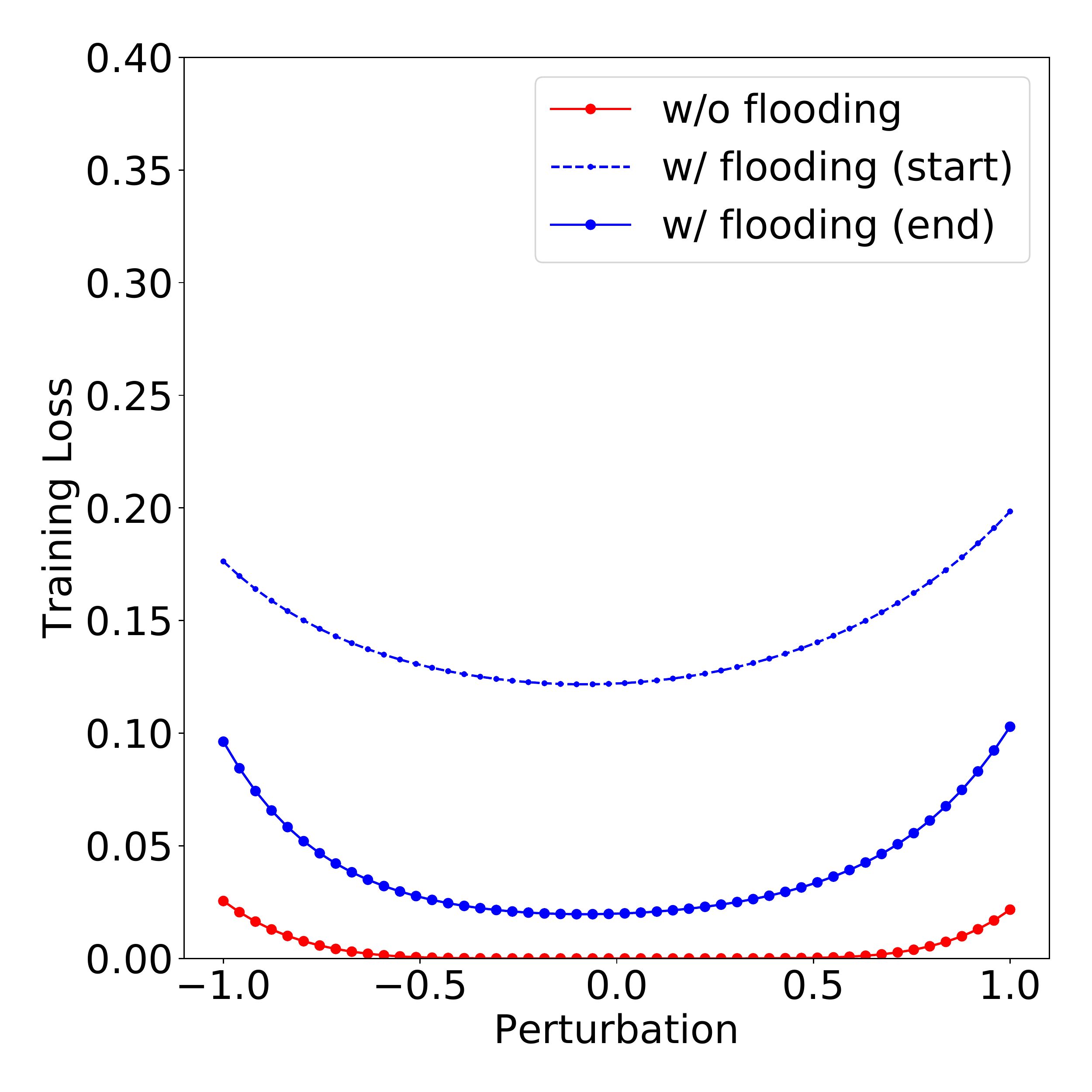}}
\subcaptionbox{MNIST (test)}{\includegraphics[width=\columnwidth*19/80]{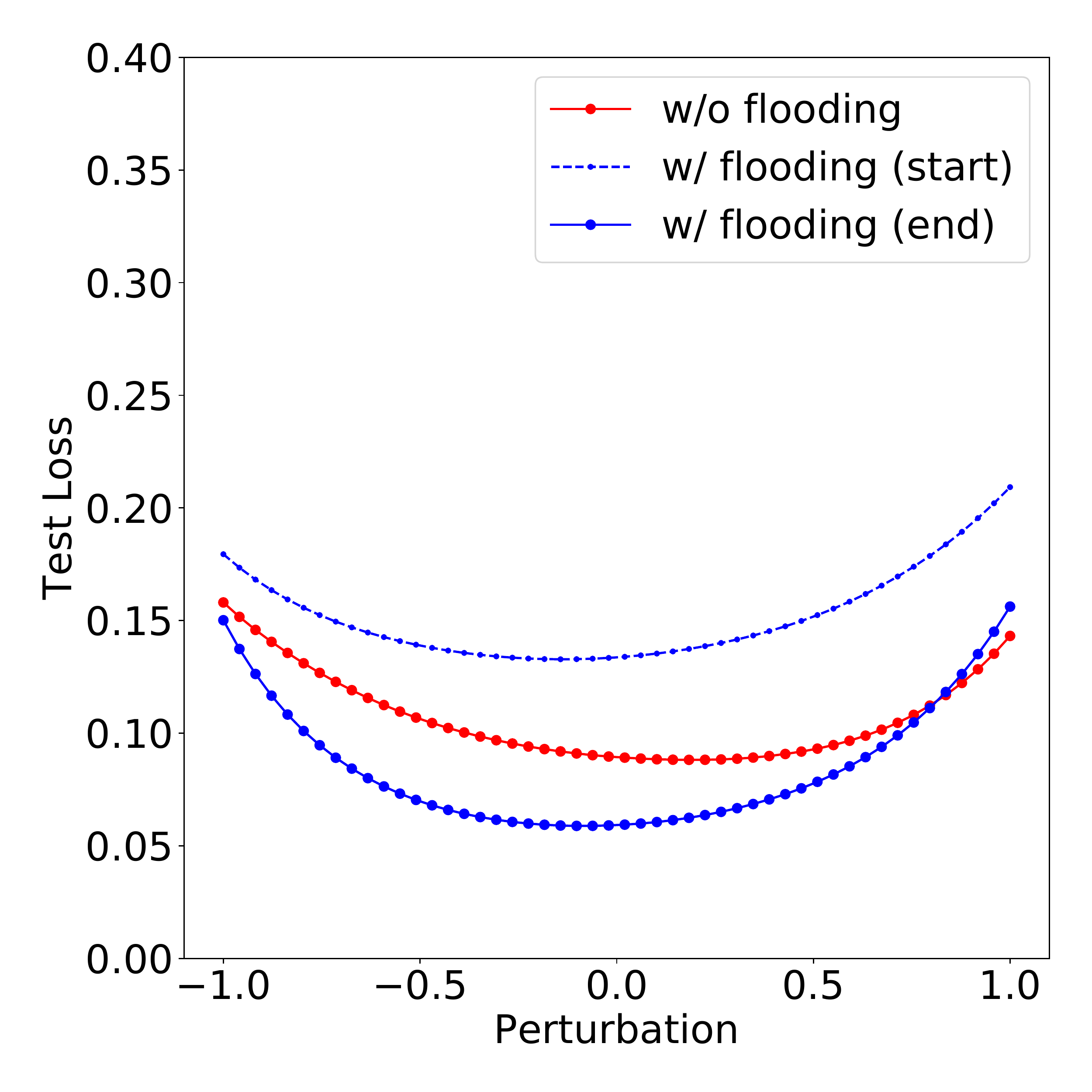}}
\subcaptionbox{KMNIST (train)}{\includegraphics[width=\columnwidth*19/80]{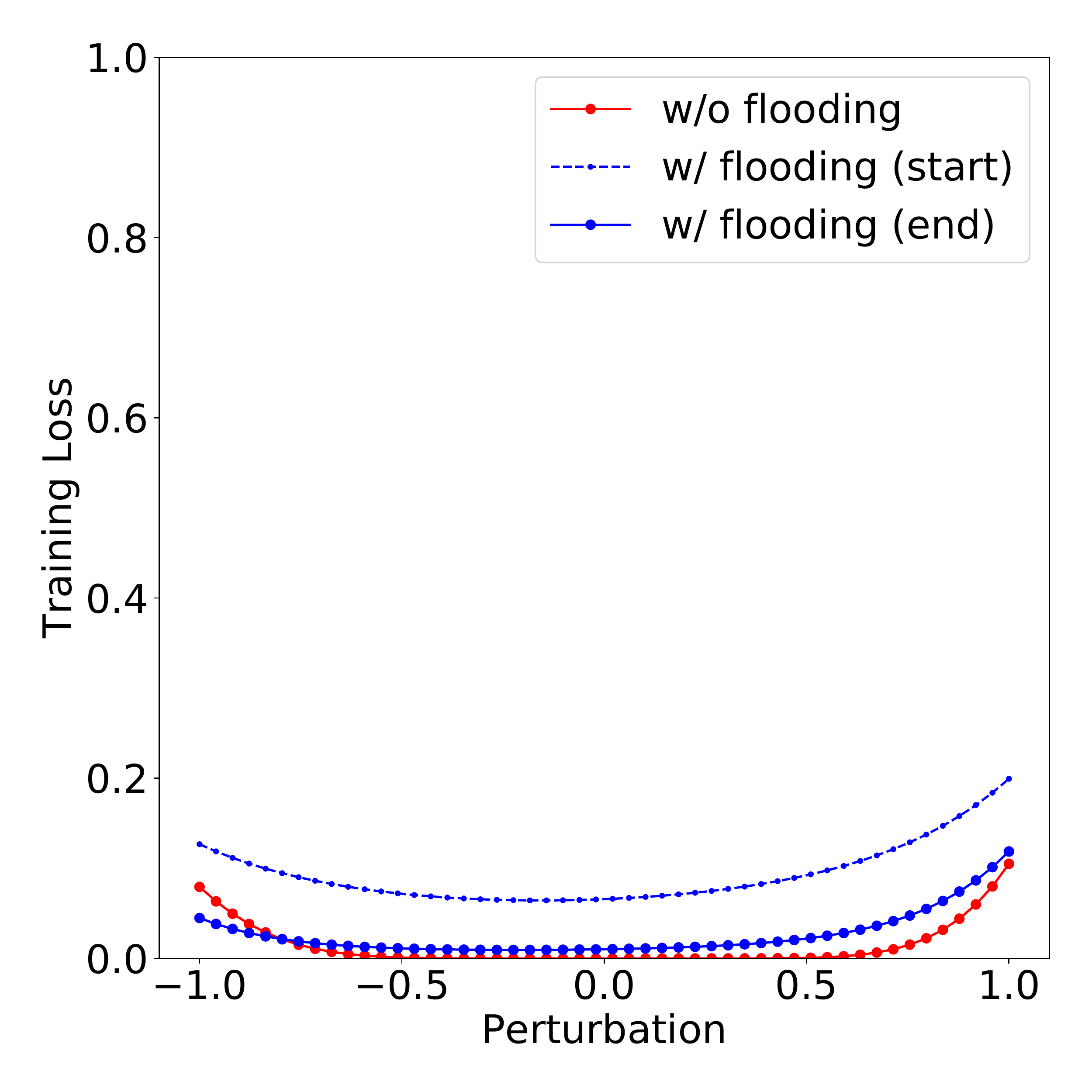}}
\subcaptionbox{KMNIST (test)}{\includegraphics[width=\columnwidth*19/80]{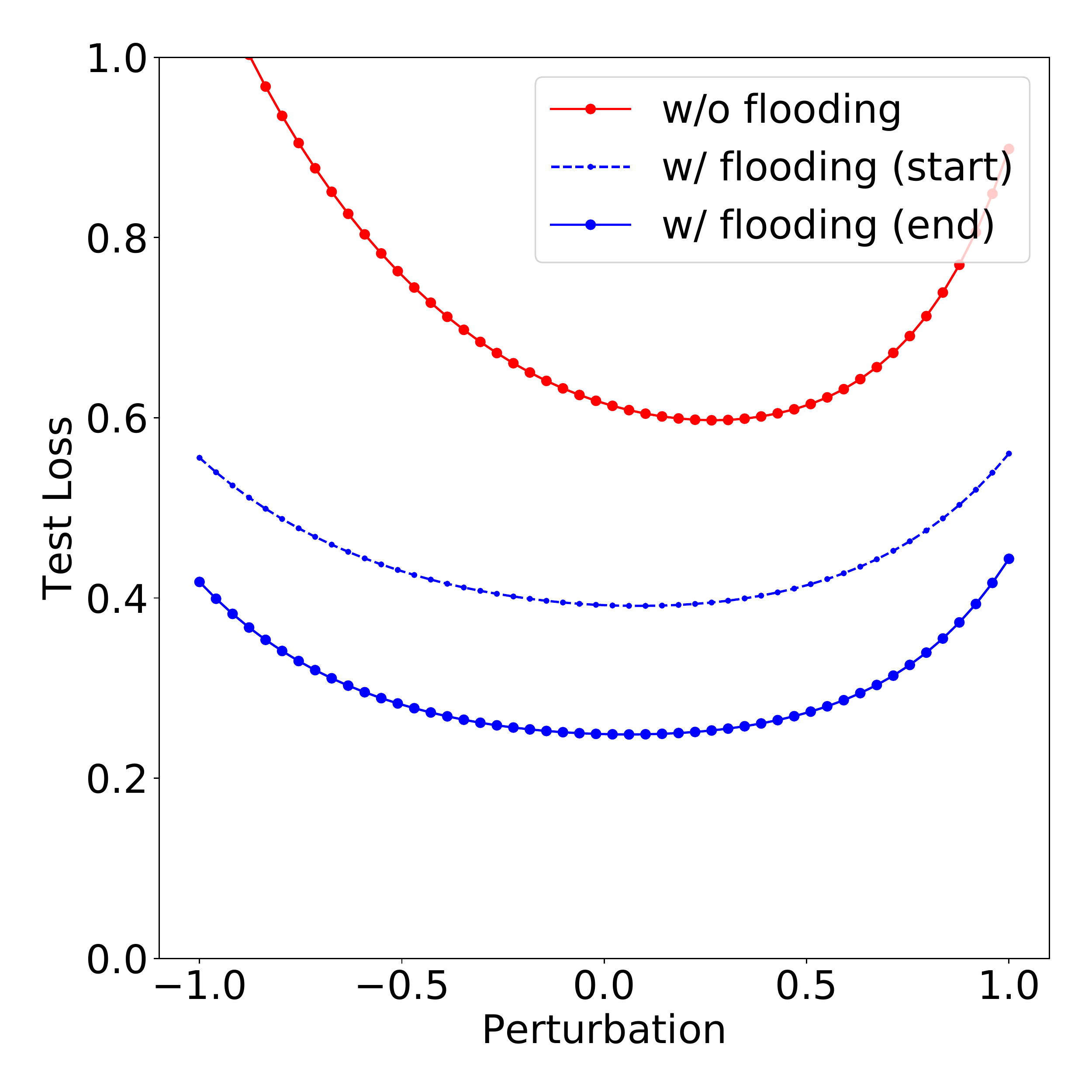}}
\subcaptionbox{SVHN (train)}{\includegraphics[width=\columnwidth*19/80]{images/flatness/flatness_SVHN_train.pdf}}
\subcaptionbox{SVHN (test)}{\includegraphics[width=\columnwidth*19/80]{images/flatness/flatness_SVHN_test.pdf}}
\subcaptionbox{CIFAR-10 (train)}{\includegraphics[width=\columnwidth*19/80]{images/flatness/flatness_C10_train.pdf}}
\subcaptionbox{CIFAR-10 (test)}{\includegraphics[width=\columnwidth*19/80]{images/flatness/flatness_C10_test.pdf}}
\subcaptionbox{CIFAR-100 (train)}{\includegraphics[width=\columnwidth*19/80]{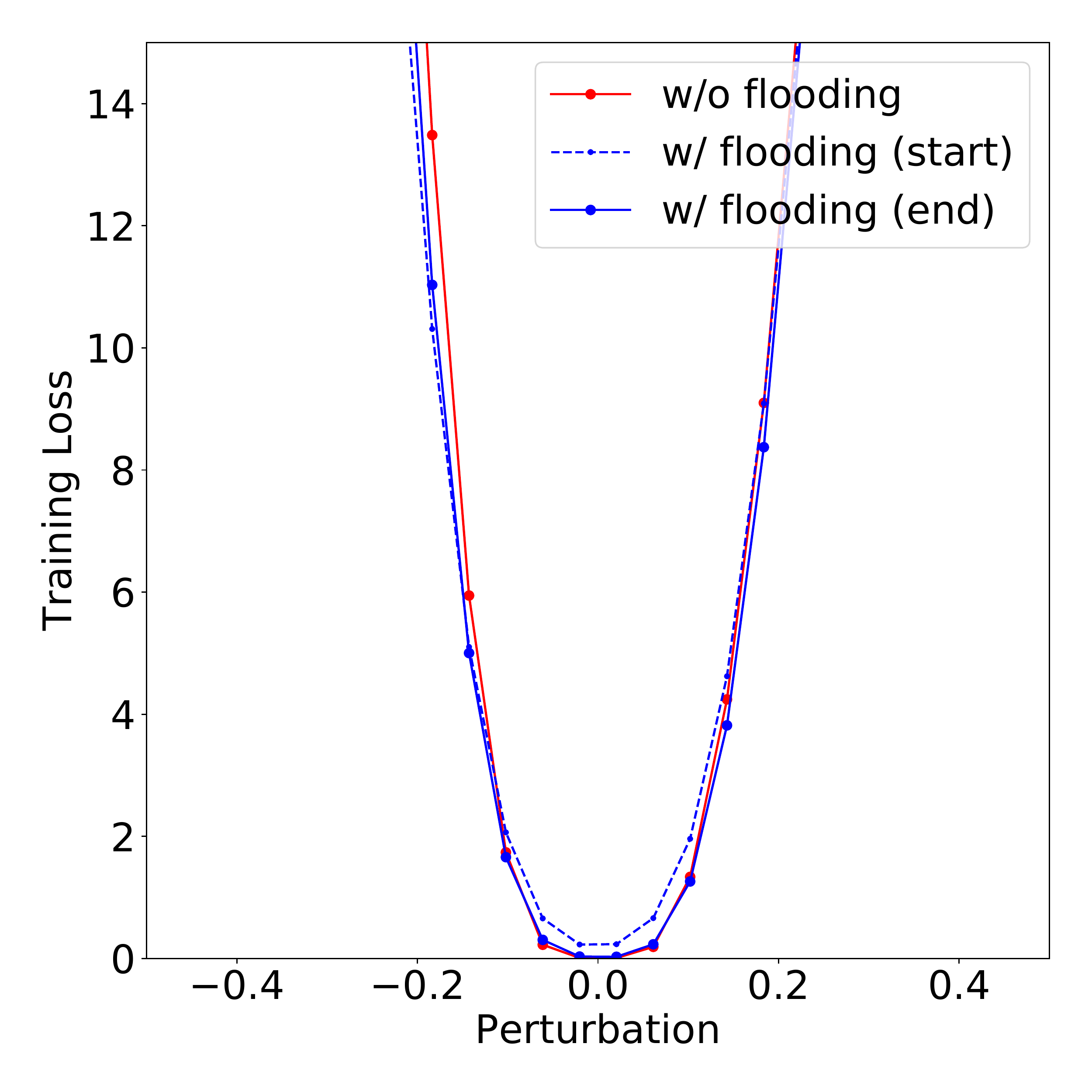}}
\subcaptionbox{CIFAR-100 (test)}{\includegraphics[width=\columnwidth*19/80]{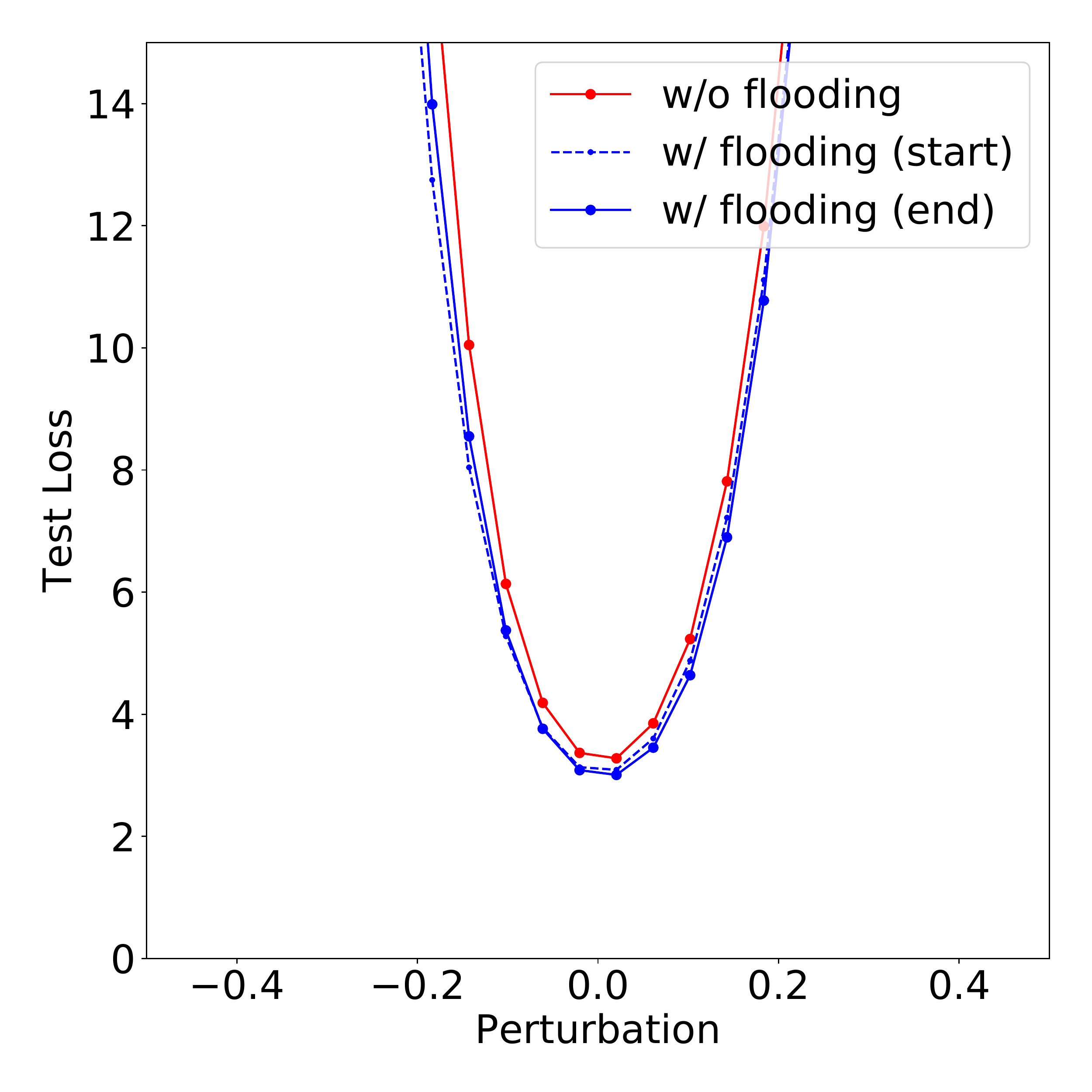}}
\caption{
One-dimensional visualization of flatness.
We visualize the training/test loss with respect to perturbation.
We depict the results for 3 models: the model when the empirical risk with respect to training data is below the flooding level for the first time during training (dotted blue), the model at the end of training with flooding (solid blue), and the model at the end of training without flooding (solid red).
}
\label{appfig:flatness}
\end{figure*}

See Fig.~\ref{appfig:performance_gradients} and Fig.~\ref{appfig:flatness} for datasets that were not included in the main paper.

\end{document}